\documentclass[10pt,twocolumn,letterpaper]{article}

\usepackage[pagenumbers]{cvpr} 
\usepackage{graphicx}
\usepackage{amsmath}
\usepackage{amssymb}
\usepackage{booktabs}
\usepackage{pifont} 
\usepackage{colortbl}
\usepackage{xcolor}
\usepackage{algorithm}
\usepackage{algpseudocode}
\usepackage{makecell}
\usepackage{multirow}
\usepackage{bm}
\usepackage{enumitem}
\usepackage{newtxtext,newtxmath}
\usepackage{algorithm}
\usepackage{algpseudocode}
\usepackage{amsmath}

\usepackage{xcolor}
\usepackage{hyperref}
\usepackage{caption}

\definecolor{evitapurple}{HTML}{7B2CBF}

\hypersetup{
    colorlinks=true,
    urlcolor=evitapurple,
    linkcolor=evitapurple,
    citecolor=evitapurple
}

\definecolor{cvprblue}{rgb}{0.21,0.49,0.74}

\def\paperID{46349} 
\def\confName{CVPR}
\def\confYear{2026}

\title{Weaving Light and Time: Unified Harmonic-Geometric Representation Learning for Dense RGB-Event Parsing}

\author{
    Chenxu Peng$^{2,1}$ \quad
    Chongtian Zhou$^{2,1}$ \quad
    Dicheng Liu$^{2,1}$ \quad
    Bowen Yin$^{2}$ \\[1mm]
    Yimian Dai$^{1,2,3}$ \quad
    Xialei Liu$^{1,2,3}$ \quad
    Ming-Ming Cheng$^{1,2,3}$ \quad
    Xiang Li$^{1,2,3}$\thanks{Corresponding author: Xiang Li.} \\[1mm]
    $^1$NKIARI, Shenzhen Futian \quad
    $^2$VCIP, CS, Nankai University \quad
    $^3$AAIS, Nankai University \\[1mm]
    {\tt\small \{cxpeng, 2311082, 2312810, bowenyin\}@mail.nankai.edu.cn} \\
    {\tt\small yimian.dai@gmail.com} \quad
    {\tt\small \{xialei, cmm, xiang.li.implus\}@nankai.edu.cn}
}

\makeatletter
\let\@oldmaketitle\@maketitle
\renewcommand{\@maketitle}{\@oldmaketitle
  \vspace{-3.2em}
  \begin{center}

    {\small
    \href{https://chaineypung.github.io/evita/}{%
        \textcolor{evitapurple}{%
          \textbf{Website: \ttfamily https://chaineypung.github.io/evita} 
        }%
    }%
    \quad \textbf{\textcolor{evitapurple}{|}} \quad
    \href{https://github.com/chaineypung/Evita}{%
        \textcolor{evitapurple}{%
          \textbf{Code: \ttfamily https://github.com/chaineypung/Evita}%
        }%
    }%
    }

    \vspace{0.8em}

    \includegraphics[width=1.0\textwidth]{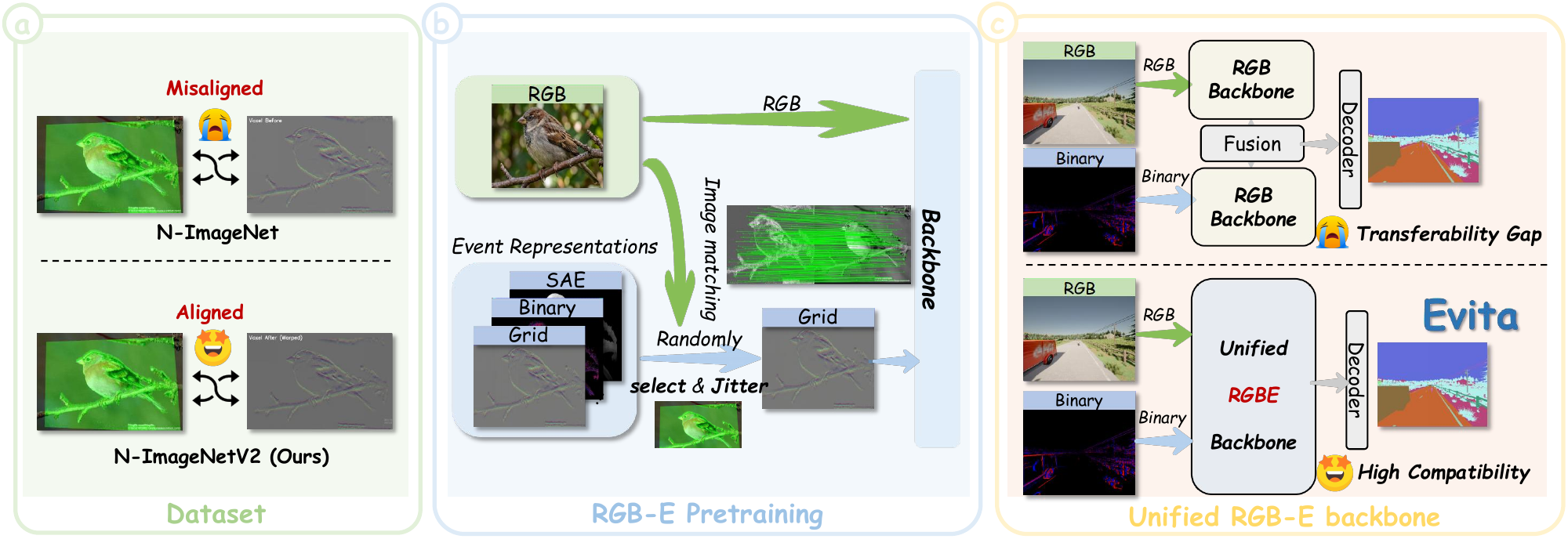}
    \captionof{figure}{Three core contributions of our work. (a) \texttt{N-ImageNetV2} resolves geometric misalignments inherent in legacy datasets to provide strictly aligned RGB and event pairs. (b) \texttt{A tailored pretraining protocol} introduces stochastic spatial jitter on well-aligned data to induce misalignment, compelling the network to autonomously learn robust invariant structural features. (c) \texttt{Evita} mitigates the prominent transfer gaps in downstream fine-tuning via a unified backbone design.}
    \label{fig:proposed_dataset}
  \end{center}
  \vspace{0.5em}
}
\makeatother

\begin{document}

\maketitle


\begin{abstract}

Fusing standard RGB frames with asynchronous event streams has emerged as a definitive paradigm for robust perception in degraded environments. Although unified backbones have recently gained traction in multi-modal vision, adapting them to the RGB-Event domain remains fundamentally challenging. Existing architectures either resort to decoupled dual encoders that double computational overhead, or adopt generic unified designs that fail to resolve implicit geometric parallax and cross-spectral aliasing under the extreme representational divide between dense intensity grids and sparse kinematic spikes. To transcend these bottlenecks, we present \texttt {Evita}, the first unified backbone specifically engineered for dedicated dense RGB-Event parsing. To achieve profound modal synergy, \texttt{Evita} explicitly embeds a suite of intrinsic co-learning modules directly into every encoder layer. Specifically, it features Geometric Parallax Rectification for adaptive spatial alignment, Harmonic Spectral Resonance for texture transfer exclusively in the complex frequency domain, and Transient Global Routing for event-driven asymmetric attention. To guarantee robust feature extraction against spatial misalignments and decouple representations from specific event encodings, we construct \texttt {N-ImageNetV2} alongside a stochastic event representation mixing pretraining protocol, empowering the network to seamlessly accommodate arbitrary event formats in downstream tasks. Extensive evaluations across the DELIVER, DDD17, and DSEC benchmarks confirm that \texttt {Evita} establishes new state-of-the-art metrics while delivering a superior accuracy-latency trade-off for real-time multimodal perception.

\end{abstract}

\section{Introduction}

\label{sec:intro}

Resilient visual perception in unconstrained environments is hindered by the limitations of standalone optical sensors~\cite{sakaridis2021acdc,chen2023explore}. While standard RGB cameras capture dense photometric details, they severely degrade under rapid motion~\cite{cho2021rethinking,zamir2021multi} and extreme illumination~\cite{guo2020zero,zamir2022restormer}. Event cameras circumvent these bottlenecks by asynchronously recording luminance changes with microsecond latency and high dynamic range~\cite{tulyakov2021time,stoffregen2020reducing}. Consequently, RGB Event fusion has emerged as a crucial paradigm for reliable dense parsing in autonomous driving and robotic perception~\cite{gehrig2021dsec,zhou2021event}.

Prevailing multimodal frameworks predominantly rely on decoupled parallel architectures~\cite{wang2022multimodal,jia2024geminifusion,li2025stitchfusion, gu2026mambaseg, bao2026re, xie2024eisnet}. These methods employ isolated backbones pretrained on static images~\cite{liu2021swin}, fusing features via complex late stage decoders. This legacy dual encoder paradigm suffers from three fundamental flaws. First, it intrinsically doubles the computational overhead and parameter count. Second, it prioritizes spatial aggregation while entirely ignoring the implicit geometric parallax caused by asynchronous sensor delays~\cite{cai2024accurate,teed2020raft}. Third, since absolute intensity and temporal contrast populate orthogonal physical spectra, simple spatial aggregation induces severe cross spectral semantic aliasing and high frequency noise insertion~\cite{weng2021event}.

While pioneering architectures like DFormer~\cite{yin2024dformer,yin2025dformerv2} and TUNI~\cite{guo2026tuni} streamline fusion in RGB Depth and RGB Thermal domains, integrating event streams into a cohesive backbone presents a fundamentally distinct challenge. Unlike synchronous dense grids, events are inherently sparse, asynchronous manifestations of relative temporal contrast. This severe representational divide between dense absolute intensity grids and continuous kinematic spikes renders conventional unified architectures ineffective. 

To transcend these bottlenecks, we present \texttt{Evita}, the first unified RGB and event backbone engineered for downstream dense prediction tasks. \texttt{Evita} abandons isolated streams by explicitly embedding cross modal interactions into every encoder layer. Specifically, each foundational block integrates three pivotal components. First, a Geometric Parallax Rectification module resolves spatial discrepancies by dynamically aligning structural boundaries via an adaptive cross modal deformable operator. Second, to overcome modality heterogeneity, a Harmonic Spectral Resonance mechanism executes cross spectral texture transfer exclusively in the complex frequency domain~\cite{li2025adaptive}, strictly avoiding spatial noise artifacts. Third, a Transient Global Routing layer formulates an asymmetric cross modal attention mechanism. By leveraging kinematic features as dynamic queries and injecting an explicit transient prior, it actively routes macroscopic photometric context from the static RGB stream.

\begin{figure}[t]
  \centering
  \includegraphics[width=\linewidth]{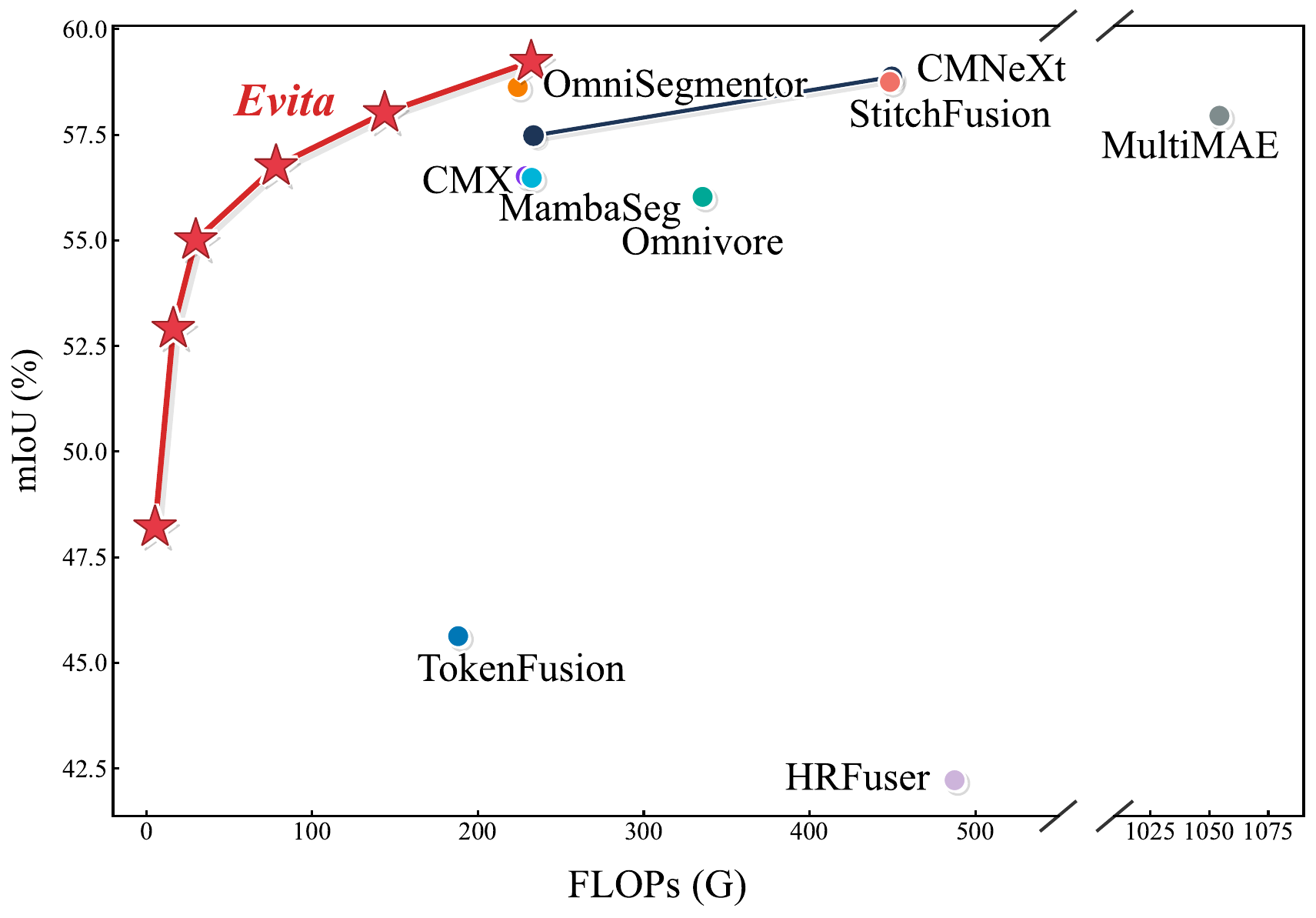}
  \caption{Performance versus computational cost on the DELIVER dataset. \texttt{Evita} achieves the new state of the art 59.57\% mIoU and the optimal accuracy computation equilibrium compared to existing methodologies.}
  \label{fig:figure8}
\end{figure}

To guarantee the extraction of highly generalized multimodal representations, we introduce a tailored cross modal pretraining paradigm. We construct \texttt{N-ImageNetV2}, a large scale dataset providing precise geometric correspondences via a semantic guided registration pipeline. Through a dual optimization loop featuring stochastic modality mixing, \texttt{Evita} systematically bridges the representation distribution shift between absolute intensity and relative contrast dynamics. Consequently, this empowers the network to seamlessly accommodate arbitrary event representations during downstream dense finetuning.

Evaluations across the DELIVER, DDD17, and DSEC benchmarks confirm that \texttt{Evita} establishes an optimal accuracy latency equilibrium, as visually summarized in Fig.~\ref{fig:figure8}. The flagship \texttt{Evita-L} achieves unprecedented state of the art metrics with drastically fewer computations, whereas the ultra lightweight \texttt{Evita-P} enables real time edge deployment. Furthermore, the learned harmonic geometric priors readily generalize to unconstrained scene parsing.

Our main contributions are summarized as follows:
\begin{itemize}[leftmargin=8pt, topsep=2pt, itemsep=1pt, parsep=0pt]
    \item \texttt{Evita}, the first unified backbone tailored for dense RGB-Event parsing, which mitigates the parameter inflation and computational redundancy of conventional decoupled paradigms.
    \item To achieve profound modal synergy, every encoder layer integrates a suite of co-learning modules: Geometric Parallax Rectification, Harmonic Spectral Resonance, and Transient Global Routing.
    \item \texttt{N-ImageNetV2} and a stochastic event representation mixing protocol are introduced, enabling the network to seamlessly accommodate arbitrary event formats during downstream finetuning.
    \item \texttt{Evita} establishes new state of the art results across multiple benchmarks, delivering an optimal accuracy latency equilibrium for real time perception.
\end{itemize}

\section{Related Work}

\noindent \textbf{Dense RGB-Event Parsing.} Standard synchronous sensors suffer from severe motion blur and dynamic range clipping under extreme conditions~\cite{zhang2022unifying,liang2024towards}. Neuromorphic event cameras circumvent these bottlenecks by asynchronously registering luminance changes with microsecond temporal resolution~\cite{gehrig2022pushing,messikommer2023data}. Since inherently sparse event streams lack static textures, dense RGB and event parsing has emerged as a compelling paradigm~\cite{jia2023event,zheng2023deep,cai2026evrwkv}. This approach leverages the strict complementarity between the photometric richness of images and the kinematic edge awareness of event data, enabling resilient scene understanding in unconstrained environments~\cite{haoyu2020learning,park2025resilient}.

To harness this multimodal synergy, methods like ESS~\cite{sun2022ess} and CMESS~\cite{xie2024cross} employ unsupervised domain adaptation to transfer structural knowledge from images, mitigating event annotation scarcity. For temporal modeling, SpikeBRGNet~\cite{long2024spike}, SLTNet~\cite{long2025sltnet}, HALSIE~\cite{das2024halsie}, and SpikingEDN~\cite{zhang2024accurate} integrate spiking neural mechanisms to efficiently capture asynchronous kinematic dynamics. Spatially, CMX~\cite{zhang2023cmx} and CMNeXt~\cite{zhang2023delivering} bridge heterogeneous semantics through multiscale cross attention interactions, while MambaSeg~\cite{gu2026mambaseg} incorporates continuous state space discretizations to elegantly map long range dependencies. Addressing physical sensor artifacts, EISNet~\cite{xie2024eisnet} leverages progressive recalibration for modality calibrated contextual synchronization, and ESC~\cite{bao2026re} explicitly recodes modal uncertainties via an edge awareness semantic concordance mechanism. Nevertheless, these predominant paradigms uniformly aggregate features extracted by two parallel pretrained backbones, rendering dual stream inference parametrically redundant and computationally inefficient. Eradicating this bottleneck, OmniSegmentor~\cite{yin2026omnisegmentor} pioneers a unified multimodal backbone, treating diverse modalities as homogeneous tokens for joint representation learning. However, this homogeneous token-mixing paradigm inherently overlooks the extreme representational divide between dense RGB grids and sparse event streams. Specifically, it heavily relies on implicit self-attention to bridge the modality gap, ignoring the physical geometric parallax and cross-spectral semantic aliasing.

\noindent \textbf{Multimodal Fusion.} The evolution of multimodal fusion is fundamentally driven by the pursuit of transferable joint representations. Early spatial representation frameworks like ESANet~\cite{seichter2021efficient} aggregated heterogeneous inputs but heavily relied on rigidly paired synthetic corpora, inevitably inducing pronounced domain gaps. With the unprecedented success of the pretrain and finetune paradigm, methodologies predominantly bifurcated into multi encoder and joint encoder designs. Multi encoder architectures, exemplified by CLIP~\cite{radford2021learning} and VATT~\cite{akbari2021vatt}, employ parallel streams to project inputs into a shared continuous space for contrastive alignment. Conversely, subsequent joint encoder frameworks such as MultiMAE~\cite{bachmann2022multimae} and Omnivore~\cite{girdhar2022omnivore} utilize unified attention mechanisms to process diverse modality tokens simultaneously, efficiently modeling early stage interactions across disparate sensory inputs. Advancing from macroscopic architectures to dense inter modality synchronization, researchers explored sophisticated feature aggregation operators. TokenFusion~\cite{wang2022multimodal} pioneered dynamic token substitution to bridge heterogeneous entities by dynamically replacing uninformative tokens. Building upon token level alignment, recent paradigms emphasize arbitrary modality integration and pixel level precision. GeminiFusion~\cite{jia2024geminifusion} implements aligned token interactions to efficiently aggregate cross modal visual cues. Concurrently, StitchFusion~\cite{li2025stitchfusion} introduces a multi directional modality adapter, enabling comprehensive cross modal information transfer and multi scale integration directly within large scale pretrained encoders. 

In this paper, we propose \texttt{Evita}, pioneering the first unified backbone engineered specifically for dense RGB-Event parsing. By explicitly rectifying geometric parallax and integrating cross spectral harmonics, it establishes a robust structural foundation. Optimized on our proposed \texttt{N-ImageNetV2} via a tailored cross modal pretraining paradigm spanning diverse event formats, \texttt{Evita} seamlessly accommodates arbitrary event representations during downstream dense finetuning, extracting pure invariant features for resilient unconstrained scene parsing.

\begin{figure*}[t]
    \vspace{-1em}
    \centering
    \includegraphics[width=1.\textwidth]{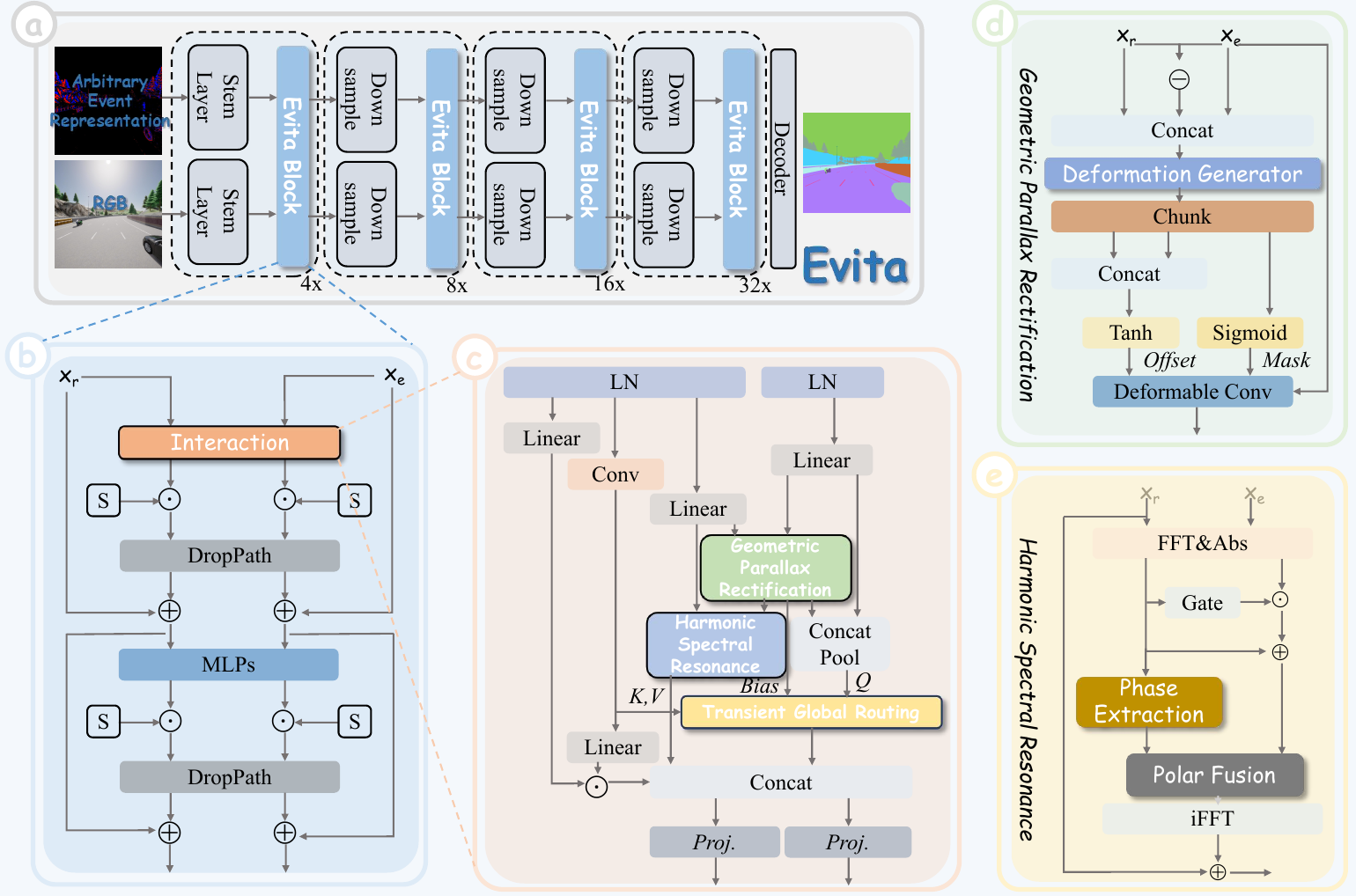}
    \caption{Overview of the proposed \texttt{Evita} framework. (a) The overall symmetric intertwined hierarchical backbone. (b) The detailed structure of the \texttt{Evita} block, enabling joint modality optimization. (c) The parallel routing mechanism integrating (d) \textit{Geometric Parallax Rectification} and (e) \textit{Harmonic Spectral Resonance} to achieve robust cross-modal feature alignment and structural frequency injection. The \textit{Transient Global Routing} layer effectively captures dynamic long-range dependencies by leveraging these structurally-aligned features as kinematic anchors.}
    \label{fig:pipeline}
    \vspace{-1em}
\end{figure*}

\section{Methodology}
\label{sec:method}

\subsection{Unified Formulation and Architecture}
Let $\mathbf{I} \in \mathbb{R}^{3 \times H \times W}$ denote a synchronous RGB frame capturing spatial radiance, and $\mathcal{E} = \{e_i\}_{i=1}^N$ represent a continuous stream of asynchronous events registration, where each tuple $e_i = (x_i, y_i, t_i, p_i)$ indexes the spatiotemporal coordinates and polarity of a localized luminance contrast perturbation~\cite{gallego2020event}. Event data can be encoded through numerous distinct representations. To alleviate the structural and temporal non-conformity between absolute intensity maps and sparse temporal derivative streams, we define a comprehensive representation repository $\mathbb{M} = \{\mathcal{R}_{\text{frame}}, \mathcal{R}_{\text{voxel}}, \mathcal{R}_{\text{sae}}, \dots\}$ which projects $\mathcal{E}$ into heterogeneous canonical tensor spaces~\cite{wang2019event, benosman2013event}. During the pretraining phase, our architecture accepts randomly sampled event representations from this repository as inputs. This stochastic sampling mechanism effectively enhances the universality of the model, compelling the network to learn robust and invariant multi-modal alignments across varying temporal and spatial event distributions.

Unlike conventional paradigms that enforce early-stage or late-stage late fusion via unconstrained heuristic networks~\cite{sun2022ess, gu2026mambaseg}, the proposed \texttt{Evita} framework instantiates a symmetric, intertwined hierarchical backbone characterized by dual-stream co-learning blocks. As illustrated in Fig.~\ref{fig:pipeline}, the inputs $\mathbf{I}$ and a randomly instantiated event representation $\mathbf{E} \in \mathbb{M}$ are projected into initial feature spaces through parallel modality-specific stem architectures. The multi-scale features are sequentially aggregated across four successive stages at resolutions of $\{1/4, 1/8, 1/16, 1/32\}$. Each constituent block within a stage implements a joint optimization of geometric alignment and cross-spectral harmonic injection, as derived in the following subsections.

\subsection{Geometric Parallax Rectification}
Owing to spatial parallax and asynchronous sensor triggered delays, the intermediate RGB features $\mathbf{X}_r \in \mathbb{R}^{C \times H \times W}$ and event features $\mathbf{X}_e \in \mathbb{R}^{C_e \times H \times W}$ typically exhibit subtle yet non-negligible geometric misalignments. To dynamically rectify this spatial discrepancy without explicit depth calibration, we formulate a non-parametric cross-modal deformable alignment operator, inspired by deformable convolutional designs~\cite{zhu2019deformable}. 

Let $\mathbf{X}'_r = \psi(\mathbf{X}_r)$ denote a linear dimension-reduced projection of the RGB stream matching the current channel capacity $C_e$ of $\mathbf{X}_e$. We define a cross-modal displacement generator function $\mathcal{G}_{\phi}$ parametrized by weights $\phi$, which operates over the concatenation of individual modalities and their directional difference tensor. Given the slight nature of the geometric misalignment, we explicitly constrain the magnitude of spatial warping via a hyperbolic tangent activator to prevent gradient instability. As illustrated in the module architecture, the intermediate features are partitioned and activated as
\begin{equation}
\begin{split}
    \mathbf{F}_x, \mathbf{F}_y, \mathbf{F}_m &= \text{Chunk}\left( \mathcal{G}_{\phi}\left( \left[ \mathbf{X}'_r \,\|\, \mathbf{X}_e \,\|\, (\mathbf{X}'_r - \mathbf{X}_e) \right] \right) \right) \\
    \mathbf{\Delta} &= \text{Tanh}\left( \left[ \mathbf{F}_x \,\|\, \mathbf{F}_y \right] \right) \in \mathbb{R}^{2K^2 \times H \times W} \\
    \mathbf{M} &= \sigma(\mathbf{F}_m) \in \mathbb{R}^{K^2 \times H \times W}
\end{split}
\end{equation}
where $\|$ signifies concatenation along the channel axis, the function $\text{Chunk}$ designates an equal channel-wise partition operator, $\mathbf{\Delta}$ represents the strictly bounded continuous horizontal and vertical coordinate offsets, and $\mathbf{M}$ is a standard modulation matrix. The geometrically rectified event representation $\hat{\mathbf{X}}_e$ at any target pixel coordinate $\mathbf{p}$ is obtained via a modulated non-local sampling operator over a local kernel support $\Omega$
\begin{equation}
\hat{\mathbf{X}}_e(\mathbf{p}) = \sum_{k \in \Omega} w_k \cdot \mathbf{X}_e\left(\mathbf{p} + \mathbf{p}_k + \mathbf{\Delta}_k(\mathbf{p})\right) \cdot \mathbf{M}_k(\mathbf{p})
\end{equation}
where $\mathbf{p}_k$ is the fixed grid offset for the $k$-th sampling position, and $w_k$ denotes the learnable kernel weights. This mechanism ensures the warp field rigorously optimizes for cross-modal structural boundaries under bounded joint guidance.

\subsection{Harmonic Spectral Resonance}
Although the geometric parallax rectification module ensures spatial layout coherence, absolute intensity representations and relative temporal contrast changes inherently populate orthogonal physical spectra. Simple spatial domain aggregation induces cross-modal aliasing and high-frequency noise insertion. To circumvent this, we implement a cross-spectral texture transfer mechanism via discrete Fourier transform operations.

Let $\mathcal{F}$ denote the Fourier transform operator. As depicted in the module architecture, the aligned representations are first projected into the complex frequency domain where the amplitude and phase components are decoupled~\cite{yang2020fda}
\begin{equation}
\mathcal{F}(\mathbf{X}'_r) = \mathcal{A}_r e^{i \Phi_r}, \quad \mathcal{F}(\hat{\mathbf{X}}_e) = \mathcal{A}_e e^{i \Phi_e}
\end{equation}
where $\mathcal{A}_r$ and $\mathcal{A}_e$ define the amplitude spectrum profiles extracted via the magnitude operations, and $\Phi_r$ corresponds to the phase structure obtained through the phase extraction branch. Noting that the phase angle preserves essential structural landmarks and global semantic topology~\cite{yang2020fda}, while the event amplitude contains invariant localized high-frequency change matrices, we introduce a parameter-controlled spectral Gate $\mathbf{G}$
\begin{equation}
\mathbf{G} = \sigma\left( \mathbf{W}_2 \cdot \text{ReLU}\left( \mathbf{W}_1 \cdot \text{AvgPool}_{H,W}(\mathcal{A}_r) \right) \right)
\end{equation}
The cross-spectral harmonic integration is realized by modulating the reference amplitude with the gated event counterpart. Subsequently, a polar fusion operation recombines this enhanced amplitude with the original phase characteristics of the radiance domain. The reconstructed feature tensor $\mathbf{X}_{\text{fused}}$ is obtained via the inverse transform $\mathcal{F}^{-1}$ coupled with a clean residual connection
\begin{equation}
\mathbf{X}_{\text{fused}} = \mathbf{X}'_r + \mathcal{F}^{-1}\left( \left( \mathcal{A}_r + \mathbf{G} \odot \mathcal{A}_e \right) \cdot e^{i \Phi_r} \right)
\end{equation}
where $\odot$ denotes the Hadamard product. This strictly aligned mathematical mapping perfectly echoes the frequency injection flow, effectively avoiding spatial noise artifacts.

\begin{table*}[t]
  \centering
  \caption{Quantitative comparison of different methods on the DELIVER dataset. The best result is highlighted in \textcolor{red}{\textbf{red}}.}
  \label{tab:comparison}
  \resizebox{0.9\textwidth}{!}{
  \begin{tabular}{lccccccc}
    \toprule
    \textbf{Method} & \textbf{Publication} & \textbf{Backbone} & \textbf{Representation} & \textbf{Input size} & \textbf{Params (M)} & \textbf{FLOPs} & \textbf{mIoU (\%)} \\
    \midrule
    HRFuser~\cite{broedermann2023hrfuser}       & ITSC'23   & HRFormer-T & EventBinary & $1024\times1024$ & 49.9  & 487.5  & 42.22 \\
    TokenFusion~\cite{wang2022multimodal}   & CVPR'22   & MiT-B2     & EventBinary & $1024\times1024$ & 26.0  & 188.05 & 45.63 \\
    MultiMAE~\cite{bachmann2022multimae}      & ECCV'22   & ViT-B      & EventBinary & $1024\times1024$ & 100.3 & 1054.3 & 57.95 \\
    Omnivore~\cite{girdhar2022omnivore}      & CVPR'22   & Swin-B     & EventBinary & $1024\times1024$ & 86.7  & 335.49 & 56.03 \\
    CMX~\cite{zhang2023cmx}           & TITS'23   & MiT-B2     & EventBinary & $1024\times1024$ & 66.6  & 228.8  & 56.52 \\
    CMNeXt~\cite{zhang2023delivering}        & CVPR'23   & MiT-B2     & EventBinary & $1024\times1024$ & 58.7  & 233.5  & 57.48 \\
    CMNeXt~\cite{zhang2023delivering}        & CVPR'23   & MiT-B4     & EventBinary & $1024\times1024$ & 116.5 & 449.6  & 58.87 \\
    MambaSeg~\cite{gu2026mambaseg}      & AAAI'25   & VMamba-T   & EventBinary & $1024\times1024$ & 25.5  & 232.45 & 56.48 \\
    OmniSegmentor~\cite{yin2026omnisegmentor} & NeurIPS'25& DFormer-L  & EventBinary & $1024\times1024$ & 39.0  & 224.0  & 58.63 \\
    StitchFusion~\cite{li2025stitchfusion}  & ACM MM'25 & MiT-B4     & EventBinary & $1024\times1024$ & 65.3  & 448.5  & 58.75 \\
    \rowcolor{blue!8} Evita-P & Ours & Ours-P & EventBinary & $1024\times1024$ & 1.3   & 5.2    & 48.21 \\
    \rowcolor{blue!8} Evita-N & Ours & Ours-N & EventBinary & $1024\times1024$ & 4.9   & 16.2   & 52.91 \\
    \rowcolor{blue!8} Evita-T & Ours & Ours-T & EventBinary & $1024\times1024$ & 6.5   & 29.8   & 55.01 \\
    \rowcolor{blue!8} Evita-S & Ours & Ours-S & EventBinary & $1024\times1024$ & 21.9  & 78.2   & 56.79 \\
    \rowcolor{blue!8} Evita-B & Ours & Ours-B & EventBinary & $1024\times1024$ & 34.3  & 143.7  & 58.02 \\
    \rowcolor{blue!8} Evita-L & Ours & Ours-L & EventBinary & $1024\times1024$ & 44.3  & 232.1  & \textcolor{red}{\textbf{59.57}} \\
    \bottomrule
  \end{tabular}
  }
\end{table*}

\begin{algorithm}[t]
\caption{Pipeline for the \texttt{N-ImageNetV2} Dataset}
\label{alg:alignment}
\begin{algorithmic}
\Require RGB $\mathbf{I}_{\text{raw}}$, event $\mathcal{E}$, extractor $\mathcal{F}_{\text{sp}}$, matcher $\mathcal{M}_{\text{lg}}$
\Ensure Registered event representation $\mathbf{E}_{\text{aligned}}$

\Statex \textbf{// Preprocessing \& Feature Extraction}
\State $\mathbf{I}_{\text{sae}} \gets \Phi_{\text{sae}}(\mathcal{E}), \ \mathbf{I}_{\text{gray}} \gets \Psi_{\text{gray}}(\mathbf{I}_{\text{raw}})$ \hfill $\triangleright$ unified mapping
\State $(\mathbf{K}_{\text{rgb}}, \mathbf{D}_{\text{rgb}}), (\mathbf{K}_{\text{evt}}, \mathbf{D}_{\text{evt}}) \gets \mathcal{F}_{\text{sp}}(\mathbf{I}_{\text{gray}}), \mathcal{F}_{\text{sp}}(\mathbf{I}_{\text{sae}})$ \hfill 

\Statex \textbf{// Iterative Auto-Alignment}
\State $\theta_{\text{conf}} \gets \theta_{\text{init}}, \ N_{\text{min}} \gets N_{\text{init}}$ \hfill $\triangleright$ initialization
\While{$\theta_{\text{conf}} \geq \theta_{\text{limit}}$} \hfill $\triangleright$ adaptive search bound
    \State $\mathbf{M} \gets \mathcal{M}_{\text{lg}}(\mathbf{D}_{\text{rgb}}, \mathbf{D}_{\text{evt}}; \theta_{\text{conf}})$ \hfill $\triangleright$ cross modal matching
    \If{$|\mathbf{M}| \geq N_{\text{min}}$}
        \State $\mathbf{H} \gets \mathcal{H}_{\text{ransac}}\big(\Pi_{\text{corr}}(\mathbf{K}_{\text{evt}}, \mathbf{K}_{\text{rgb}} \mid \mathbf{M})\big)$ 
        \If{$\epsilon_{\text{min}} < |\det(\mathbf{H})| < \epsilon_{\text{max}}$}
            \State \Return $\mathcal{T}_{\text{warp}}(\mathbf{I}_{\text{sae}}, \mathbf{H})$ \hfill $\triangleright$ automated return
        \EndIf
    \EndIf
    \State $\theta_{\text{conf}} \gets \theta_{\text{conf}} - \Delta\theta, \ N_{\text{min}} \gets \max(N_{\text{min}} - \Delta N, \ 5)$ \hfill 
\EndWhile

\Statex \textbf{// Human-in-the-loop Fallback}
\State $\mathbf{H}_{\text{manual}} \gets \mathcal{H}_{\text{dlt}}\big(\Omega_{\text{manual}}(\mathbf{I}_{\text{raw}}, \mathbf{I}_{\text{sae}})\big)$ \hfill
\State \Return $\mathcal{T}_{\text{warp}}(\mathbf{I}_{\text{sae}}, \mathbf{H}_{\text{manual}})$ \hfill $\triangleright$ calibrated return
\end{algorithmic}
\end{algorithm}

\subsection{Transient Global Routing}
As illustrated in the comprehensive block architecture, rather than adopting a conventional sequential pipeline, the \texttt{Evita} block instantiates a sophisticated parallel routing mechanism. The network maps dynamic long-range dependencies through a transient global routing layer which operates in strictly parallel with the frequency injection branch.

To formulate a genuine cross-modal attention mechanism, the keys and values are extracted from the normalized RGB stream via a spatial convolutional mapping $\mathbf{K}, \mathbf{V} = \text{Conv}(\text{LN}(\mathbf{X}_r))$. Conversely, the queries are constructed entirely from the kinematic event domain. The geometrically aligned event features from the alignment module and the linearly projected event representations are aggregated via a joint concatenation and pooling operator to generate the query matrix $\mathbf{Q} = \text{Pool}([\hat{\mathbf{X}}_e \,\|\, \mathbf{W}_e\text{LN}(\mathbf{X}_e)])$.

Furthermore, an explicit transient prior tensor $\mathbf{B}_{\text{evt}}$ is computed directly from the aligned event stream $\hat{\mathbf{X}}_e$ prior to any frequency modifications. The modulated cross-attention operator is rigorously defined as
\begin{equation}
\text{TGR}(\mathbf{Q}, \mathbf{K}, \mathbf{V}) = \text{Softmax}\left( \frac{\mathbf{Q}\mathbf{K}^T}{\sqrt{d_k}} + \mathbf{B}_{\text{evt}} \right) \mathbf{V}
\end{equation}
Ultimately, the output of this structured routing module is concatenated with the frequency-injected features and a gated residual RGB branch. The aggregated tensor is then passed through parallel linear projections to yield the updated dual-stream representations for the subsequent stage.

\subsection{RGB-E Pretraining}

\noindent \textbf{N-ImageNetV2.} We present \texttt{N-ImageNetV2}, a large-scale multimodal benchmark designed to eliminate the severe geometric parallax prevalent in legacy datasets. Built upon the immense diversity of N-ImageNet~\cite{kim2021n}, our dataset provides over 1.2 million strictly aligned RGB and event pairs spanning 1K semantic categories. As outlined in Algorithm~\ref{alg:alignment}, the proposed registration pipeline initiates by mapping the sparse events $\mathcal{E}$ into a Surface of Active Events (SAE)~\cite{benosman2013event} to establish a reliable geometric proxy:
\begin{equation}
\mathcal{R}_{\text{sae}}(x, y) = \max \{t_i \mid (x_i, y_i) = (x, y), t_i \leq t_{\text{ref}}\}
\end{equation}

Dense structural keypoints are subsequently extracted from both the grayscale RGB and SAE representations via SuperPoint~\cite{detone2018superpoint}, followed by cross modal correspondence matching using LightGlue~\cite{lindenberger2023lightglue}. To maximize the automated registration yield, we design an iterative adaptive relaxation strategy. This mechanism progressively decays the confidence thresholds and inlier bounds until RANSAC~\cite{wei2023generalized} successfully estimates a robust homography matrix $\mathbf{H}$ to warp the event streams into the static RGB coordinate space. For extreme scenarios that fail the automated geometric validation, we trigger a human-in-the-loop fallback. By coupling precise manual point pairing with the direct linear transform, this calibration phase resolves all hard cases and guarantees absolute spatial alignment across the entire dataset.

\noindent \textbf{Pretraining Strategy.} Rather than enforcing rigid alignment, we adopt a hybrid training objective to enhance spatial robustness. During training, we randomly apply synthetic geometric perturbations to the aligned pairs from \texttt{N-ImageNetV2}, forcing the network to oscillate between inherently aligned and dynamically misaligned data states. This adaptive strategy prevents overfitting to the alignment module itself, forcing the geometric parallax rectification module to master dynamic spatial discrepancy estimation while maintaining inherent structural robustness. Let $\mathcal{P}$ denote the set of pixel coordinates where confident matches are registered. We append a geometric coherence penalty to the global classification objective
\begin{equation}
\mathcal{L}_{\text{total}} = \mathcal{L}_{\text{cls}} + \lambda \cdot \mathbb{1}_{\{\mathcal{P} \neq \emptyset\}} \sum_{\mathbf{u} \in \mathcal{P}} \left\| \mathbf{\Delta}(\mathbf{u}) - \left( \frac{\mathbf{H}_{\text{gt}}\tilde{\mathbf{u}}}{(\mathbf{H}_{\text{gt}}\tilde{\mathbf{u}})_3} - \mathbf{u} \right) \right\|_1
\end{equation}
where $\tilde{\mathbf{u}} = [x, y, 1]^T$ indicates the homogeneous coordinate vector of pixel $\mathbf{u}$, and the third subscript isolates the scaling component for perspective projection normalization. The indicator function restricts the alignment update to valid image and event pairs, ensuring optimization stability.

\section{Experiments}

\begin{table*}[t]

  \centering

  \caption{Quantitative comparison of different methods on DDD17 and DSEC datasets. The best results are highlighted in \textcolor{red}{\textbf{red}}.}

  \label{tab:comprehensive_comparison}


  \resizebox{\textwidth}{!}{

  \begin{tabular}{lccccccccccccc}

    \toprule

    \multirow{2}{*}{\textbf{Method}} & \multirow{2}{*}{\textbf{Publication}} & \multirow{2}{*}{\textbf{Backbone}} & \multirow{2}{*}{\textbf{Params (M)}} & \multirow{2}{*}{\textbf{Modality}} & \multirow{2}{*}{\textbf{Representation}} & \multicolumn{4}{c}{\textbf{DDD17}} & \multicolumn{4}{c}{\textbf{DSEC}} \\

    \cmidrule(lr){7-10} \cmidrule(lr){11-14}

    & & & & & & \textbf{Input size} & \textbf{FLOPs (G)} & \textbf{mIoU (\%)} & \textbf{Acc (\%)} & \textbf{Input size} & \textbf{FLOPs (G)} & \textbf{mIoU (\%)} & \textbf{Acc (\%)} \\

    \midrule

    \multicolumn{14}{l}{\textbf{Single Modal Method}} \\

    \midrule

    SegFormer~\cite{xie2021segformer}       & NeurIPS'21  & MiT-B2     & 27.5  & Image       & RGB          & $200\times346$ & 16.0  & 71.05 & 95.73 & $440\times640$ & 67.6  & 71.99 & 94.97 \\

    SegNeXt~\cite{guo2022segnext}         & NeurIPS'22  & SegNeXt-B  & 27.6  & Image       & RGB          & $200\times346$ & 10.2  & 71.46 & 95.97 & $440\times640$ & 40.4  & 71.55 & 94.89 \\

    EV-SegNet~\cite{alonso2019ev}       & CVPR'19     & Xception   & 34.6  & Event       & 6-Channel    & $200\times346$ & 8.0   & 54.81 & 89.76 & $440\times640$ & 30.6  & 51.76 & 88.61 \\

    ESS~\cite{sun2022ess}             & ECCV'22     & E2VID      & 47.2  & Event       & Voxel Grid   & $200\times346$ & 51.8  & 61.37 & 91.08 & $440\times640$ & 202.6 & 51.57 & 89.25 \\

    \midrule

    \multicolumn{14}{l}{\textbf{RGB-X Method}} \\

    \midrule

    TokenFusion~\cite{wang2022multimodal}     & CVPR'22     & MiT-B2     & 28.6  & Image-Event & Voxel Grid   & $200\times346$ & 32.6  & 67.96 & 94.77 & $440\times640$ & 137.4 & 70.67 & 95.20 \\

    MultiMAE~\cite{bachmann2022multimae}        & ECCV'22     & ViT-B      & 89.7  & Image-Event & Voxel Grid   & $200\times346$ & 56.1  & 63.31 & 93.33 & $440\times640$ & 291.5 & 67.11 & 94.35 \\

    HRFuser~\cite{broedermann2023hrfuser}         & ITSC'23     & HRFormer-T & 12.8  & Image-Event & Voxel Grid   & $200\times346$ & 17.6  & 73.17 & 95.41 & $440\times640$ & 70.0  & 59.66 & 92.76 \\

    CMX~\cite{zhang2023cmx}             & TITS'23     & MiT-B2     & 66.6  & Image-Event & Voxel Grid   & $200\times346$ & 15.7  & 77.64 & 96.46 & $440\times640$ & 62.1  & 73.88 & 95.51 \\

    CMX~\cite{zhang2023cmx}             & TITS'23     & MiT-B3     & 106.3 & Image-Event & Voxel Grid   & $200\times346$ & 23.8  & 76.83 & 96.50 & $440\times640$ & 94.1  & 73.24 & 95.40 \\

    CMX~\cite{zhang2023cmx}             & TITS'23     & MiT-B4     & 140.0 & Image-Event & Voxel Grid   & $200\times346$ & 31.5  & 77.00 & 96.36 & $440\times640$ & 124.8 & 74.06 & 95.54 \\

    CMNeXt~\cite{zhang2023delivering}          & CVPR'23     & MiT-B2     & 58.7  & Image-Event & Voxel Grid   & $200\times346$ & 16.0  & 76.86 & 96.47 & $440\times640$ & 63.4  & 73.42 & 95.48 \\

    CMNeXt~\cite{zhang2023delivering}          & CVPR'23     & MiT-B4     & 116.6 & Image-Event & Voxel Grid   & $200\times346$ & 31.0  & 77.70 & 96.48 & $440\times640$ & 122.3 & 73.84 & 95.49 \\

    CMNeXt~\cite{zhang2023delivering}          & CVPR'23     & MiT-B5     & 149.0 & Image-Event & Voxel Grid   & $200\times346$ & 37.8  & 77.09 & 96.47 & $440\times640$ & 149.3 & 74.22 & 95.59 \\

    GeminiFusion~\cite{jia2024geminifusion}    & ICML'24     & MiT-B2     & 31.8  & Image-Event & Voxel Grid   & $200\times346$ & 33.6  & 71.15 & 96.15 & $440\times640$ & 141.4 & 71.45 & 95.31 \\

    StitchFusion~\cite{li2025stitchfusion}    & ACM MM'25   & MiT-B2     & 30.0  & Image-Event & Voxel Grid   & $200\times346$ & 20.8  & 76.83 & 96.30 & $440\times640$ & 89.7  & 73.63 & 95.43 \\

    StitchFusion~\cite{li2025stitchfusion}    & ACM MM'25   & MiT-B3     & 47.5  & Image-Event & Voxel Grid   & $200\times346$ & 28.2  & 76.58 & 96.54 & $440\times640$ & 112.1 & 74.47 & 95.57 \\

    StitchFusion~\cite{li2025stitchfusion}    & ACM MM'25   & MiT-B4     & 64.4  & Image-Event & Voxel Grid   & $200\times346$ & 36.0  & 77.41 & 96.50 & $440\times640$ & 143.1 & 73.69 & 95.44 \\

    OmniSegmentor~\cite{yin2026omnisegmentor}   & NeurIPS'25  & DFormer-L  & 40.9  & Image-Event & Voxel Grid   & $200\times346$ & 25.3  & 79.34 & 96.54 & $440\times640$ & 100.7 & 75.31 & 95.74 \\

    \midrule

    \multicolumn{14}{l}{\textbf{RGB-E Method}} \\

    \midrule

    ESS~\cite{sun2022ess}             & ECCV'22     & E2VID      & 52.7  & Image-Event & Voxel Grid   & $200\times346$ & 20.0  & 60.43 & 90.37 & $440\times640$ & 77.4  & 53.29 & 89.37 \\

    EDCNet~\cite{zhang2021exploring}          & TITS'22     & ResNet-18  & 23.1  & Image-Event & Voxel Grid   & $256\times352$ & 21.3  & 61.99 & 93.80 & $480\times640$ & 72.5  & 56.75 & 92.39 \\

    HALSIE~\cite{das2024halsie}          & WACV'24     & Customed   & 5.0   & Image-Event & Voxel Grid   & $192\times192$ & 4.9   & 60.66 & 92.50 & $440\times640$ & 37.7  & 52.43 & 89.01 \\

    SE-Adapter~\cite{yao2024sam}      & ICRA'24     & ViT-B      & 88.1  & Image-Event & MSP          & $256\times256$ & 62.3  & 69.06 & 95.32 & $480\times640$ & 267.4 & 69.77 & 93.58 \\

    EISNet~\cite{xie2024eisnet}          & TMM'24      & MiT-B2     & 34.4  & Image-Event & AET          & $200\times346$ & 16.8  & 75.03 & 96.04 & $440\times640$ & 67.3  & 73.07 & 95.12 \\

    EISNet~\cite{xie2024eisnet}          & TMM'24      & MiT-B3     & 95.0  & Image-Event & AET          & $200\times346$ & 29.3  & 76.14 & 96.14 & $440\times640$ & 116.6 & 73.33 & 95.07 \\

    EISNet~\cite{xie2024eisnet}          & TMM'24      & MiT-B4     & 128.5 & Image-Event & AET          & $200\times346$ & 37.0  & 76.06 & 95.89 & $440\times640$ & 147.2 & 72.56 & 95.05 \\

    Hybrid-Seg~\cite{li2025efficient}      & AAAI'25     & ERViT-T    & 2.1   & Image-Event & Voxel Grid   & $200\times346$ & 4.0   & 67.31 & 95.07 & $440\times640$ & 14.4  & 66.57 & 94.27 \\

    ESC~\cite{bao2026re}             & NeurIPS'25  & MiT-B2     & 56.9  & Image-Event & Voxel Grid   & $200\times346$ & 18.3  & 76.81 & 96.35 & $440\times640$ & 95.1  & 73.55 & 95.49 \\

    MambaSeg~\cite{gu2026mambaseg}        & AAAI'25     & VMamba-T   & 25.4  & Image-Event & Voxel Grid   & $200\times346$ & 33.8  & 77.56 & 96.33 & $440\times640$ & 159.5 & 75.10 & 95.71 \\

    \rowcolor{blue!8} Evita-Pico & Ours       & Ours-P     & 1.3   & Image-Event & Voxel Grid   & $200\times346$ & 0.6   & 71.26 & 95.78 & $440\times640$ & 2.6   & 65.55 & 94.12 \\

    \rowcolor{blue!8} Evita-Nano & Ours       & Ours-N     & 4.9   & Image-Event & Voxel Grid   & $200\times346$ & 2.3   & 74.84 & 96.17 & $440\times640$ & 9.2   & 70.09 & 94.99 \\

    \rowcolor{blue!8} Evita-Tiny & Ours       & Ours-T     & 6.5   & Image-Event & Voxel Grid   & $200\times346$ & 3.2   & 77.04 & 96.45 & $440\times640$ & 12.9  & 73.90 & 95.55 \\

    \rowcolor{blue!8} Evita-Small& Ours       & Ours-S     & 21.9  & Image-Event & Voxel Grid   & $200\times346$ & 10.3  & 78.56 & 96.72 & $440\times640$ & 40.9  & 75.07 & 95.70 \\

    \rowcolor{blue!8} Evita-Base & Ours       & Ours-B     & 34.3  & Image-Event & Voxel Grid   & $200\times346$ & 14.7  & 79.11 & \textcolor{red}{\textbf{96.73}} & $440\times640$ & 58.5  & 76.08 & 95.86 \\

    \rowcolor{blue!8} Evita-Large& Ours       & Ours-L     & 44.3  & Image-Event & Voxel Grid   & $200\times346$ & 20.8  & \textcolor{red}{\textbf{80.12}} & 96.68 & $440\times640$ & 82.7  & \textcolor{red}{\textbf{76.80}} & \textcolor{red}{\textbf{95.97}} \\

    \bottomrule

  \end{tabular}

  }

\end{table*}

\subsection{Implementation Details}

\textbf{Pretraining settings.} We execute the pretraining process on the ImageNet-1K and our proposed \texttt{N-ImageNetV2} datasets utilizing 8×NVIDIA 3090 GPUs. To endow the \texttt{Evita} backbone with robust spatial adaptability, we introduce a stochastic alignment protocol. During training, the RGB and event representations are explicitly aligned with a 60 percent probability, while the remaining 40 percent are intentionally left misaligned. This dynamic training environment compels the model to actively learn how to perform geometric registration rather than simply memorizing static spatial correspondences. The multimodal inputs are resized to a spatial resolution of 224 by 224. Data augmentation strategies altering color and illumination are applied exclusively to the RGB images. Conversely, spatial augmentations such as random rotation and cropping are applied synchronously to both domains to strictly preserve their structural coherence. We utilize the standard cross entropy loss as our optimization objective and train the network for 300 epochs. Optimization is performed via AdamW with a learning rate of 0.001, a weight decay of 0.05, and a global batch size of 1024. For structural regularization, the drop path rates are specifically scaled to 0.0, 0.05, 0.1, 0.1, 0.15, and 0.2 across the P, N, T, S, B, and L variants of \texttt{Evita} respectively. Comprehensive architectural specifics are available in the supplementary material.

\noindent \textbf{Datasets and setting for finetuning.} To rigorously evaluate the generalization capability of our unified architecture across diverse environmental conditions encompassing adverse weather, low illumination, and clear driving scenes, we conduct finetuning experiments on the DELIVER~\cite{zhang2023delivering}, DDD17~\cite{alonso2019ev}, and DSEC~\cite{sun2022ess} benchmarks evaluated via the mean Intersection over Union metric. For the DELIVER dataset, we strictly adhere to the established CMNeXt protocol where the multimodal images are augmented through random resizing with scale ratios between 0.5 and 2.0, random horizontal flipping, random color jittering, random gaussian blurring, and subsequent random cropping to a spatial resolution of 1024 by 1024 pixels. Regarding the DDD17 and DSEC benchmarks, our configuration aligns with recent event based segmentation methodologies~\cite{gu2026mambaseg}. We discretize the raw asynchronous streams into ten bin voxel grids utilizing fixed 50 millisecond temporal windows and precise aggregates of 100k events respectively. To mitigate overfitting on these two event specific datasets, standard spatial augmentations including random cropping, horizontal flipping, and stochastic scale variations are applied. The network optimization is driven by the AdamW optimizer paired with a standard cross entropy objective function, allocating an initial learning rate of 2e-4 and a batch size of 12 for DDD17, while adjusting to a learning rate of 6e-5 with a batch size of 4 for DSEC.

\begin{figure*}[t]
    \centering
    \includegraphics[width=\linewidth]{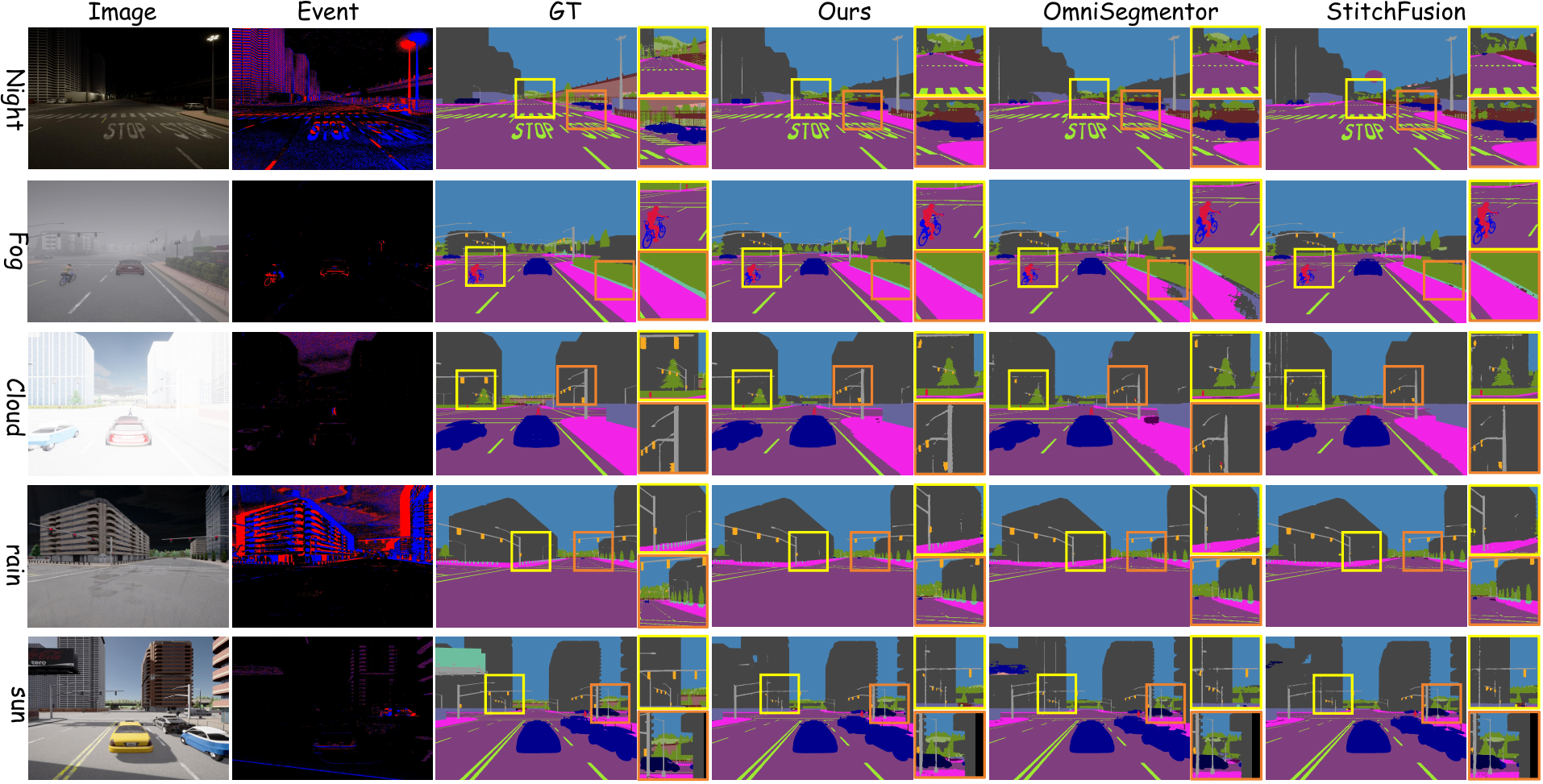}
    \vspace{-0.5em}
    \caption{Qualitative comparison on the DELIVER dataset under diverse adverse weather conditions. \texttt{Evita} exhibits superior robustness in reconstructing fine-grained structures and maintaining semantic consistency, particularly in low-light and high-occlusion scenarios, outperforming competing architectures.}
    \label{fig:consensus_vis}
\end{figure*}

\subsection{Comparison with State-of-the-art Methods}

We compare our proposed \texttt{Evita} architecture against seventeen recent multimodal semantic segmentation methodologies across the DELIVER, DDD17, and DSEC benchmarks. As detailed in Table~\ref{tab:comparison} for the DELIVER dataset, the flagship \texttt{Evita-L} achieves a new state of the art performance of 59.57 percent mIoU. Notably, it surpasses the formidable CMNeXt-B4 by 0.36\% while utilizing merely 38 percent of its parameters and approximately halving the computational FLOPs. This performance gain is consistently reflected in our qualitative evaluations, where \texttt{Evita} exhibits superior structural clarity and semantic coherence across challenging scenarios, as demonstrated in Fig.~\ref{fig:consensus_vis}. Furthermore, our scaled down variant \texttt{Evita-S} delivers highly competitive accuracy at 56.76\% with only 21.9M parameters, establishing a strictly superior accuracy computation equilibrium compared to massive legacy architectures like MultiMAE and Omnivore.

\begin{table}[t]
  \centering
  \caption{Ablation study on the effect of different pretraining datasets. Models are evaluated using the mIoU metric on DDD17.}
  \label{tab:pretrained_dataset}
  \resizebox{\columnwidth}{!}{
  \begin{tabular}{cccc}
    \toprule
    \multicolumn{3}{c}{Pretrained Dataset} & \multirow{2}{*}{mIoU (\%)} \\
    \cmidrule(lr){1-3}
    ImageNet-1K & N-ImageNet & N-ImageNetV2 (Ours) & \\
    \midrule
    \checkmark & & & 76.71 \\
    \checkmark & \checkmark & & 77.82 {\scriptsize\textcolor{blue}{(+1.11)}} \\
    \rowcolor{blue!8} \checkmark & & \checkmark & \textbf{78.25} {\scriptsize\textcolor{blue}{(+1.54)}} \\
    \bottomrule
  \end{tabular}
  }
\end{table}

Table~\ref{tab:comprehensive_comparison} extends our evaluation to the DDD17 and DSEC driving benchmarks. \texttt{Evita-L} establishes new state-of-the-art results with 80.12\% and 76.80\% mIoU, profoundly eclipsing leading frameworks like OmniSegmentor and MambaSeg. Critically, \texttt{Evita-L} requires only 82.7 GFLOPs on DSEC. This is nearly half the computational burden of MambaSeg with 159.5 GFLOPs, yet it yields a substantial 2.0\% absolute accuracy improvement. Furthermore, the ultra-lightweight \texttt{Evita-P} and \texttt{Evita-N} deliver robust baselines with negligible overhead for edge deployment. These empirical gains confirm that our unified harmonic-geometric paradigm accurately extracts cross-modal features, entirely circumventing the parameter inflation of traditional decoupled networks.

\section{Ablation Study and Analysis}

\begin{table}[t]
  \centering
  \caption{Cross-architecture generalization of our RGB-E pretraining protocol. For CMX and CMNeXt: $^\ast$ denotes training from scratch; $^\S$ denotes initialization with RGB priors. For Dformer: $^\dag$ denotes direct RGB-E pretraining; $^\ddag$ denotes initialization with native RGB-D priors.}
  \label{tab:model_comparison}
  \resizebox{0.48\textwidth}{!}{
  \begin{tabular}{lccccc}
    \toprule
    Model & Params & DDD17 & FLOPs & DSEC & FLOPs\\
    \midrule
    CMX-B2$^\ast$ & 66.6M & 71.49 & 15.7G & 67.37 & 62.1G\\
    CMX-B2$^\S$ & 66.6M & 78.17 & 15.7G & 74.23 & 62.1G\\
    CMNeXt-B2$^\ast$ & 58.7M & 72.08 & 16.0G & 67.85 & 63.4G\\
    CMNeXt-B2$^\S$ & 58.7M & 78.62 & 16.0G & 74.58 & 63.4G\\
    OminiSegmentor-L$^\dag$ & 40.9M & 73.31 & 25.3G & 70.45 & 100.7G\\
    OminiSegmentor-L$^\ddag$ & 40.9M & 79.46 & 25.3G & 75.89 & 100.7G\\
    \rowcolor{blue!8} Evita-L & 44.3M & \textbf{80.12} & 20.8G & \textbf{76.80} & 82.7G\\
    \bottomrule
  \end{tabular}
  }
\end{table}

\begin{figure}[t]
  \centering
  
  
  
  \begin{subfigure}{\linewidth}
    \centering
    \includegraphics[width=0.9\linewidth]{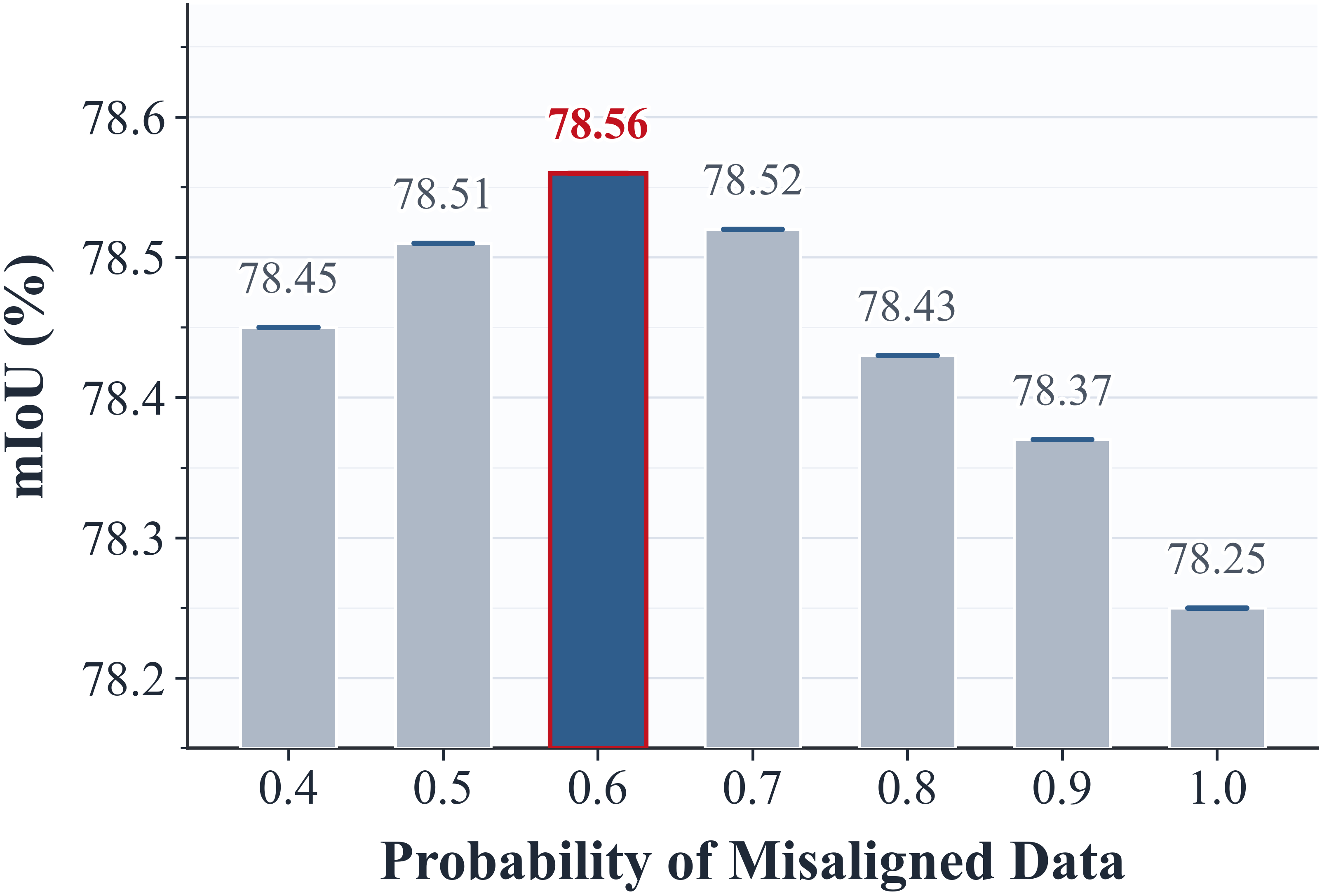}
    \label{fig:analysis_misalign}
  \end{subfigure}
  
  \caption{The ablation study on stochastic misalignment probability peaks at 0.6.}
  \label{fig:comprehensive_analysis}
\end{figure}

\begin{figure}[b]
    \centering
    \includegraphics[width=\linewidth]{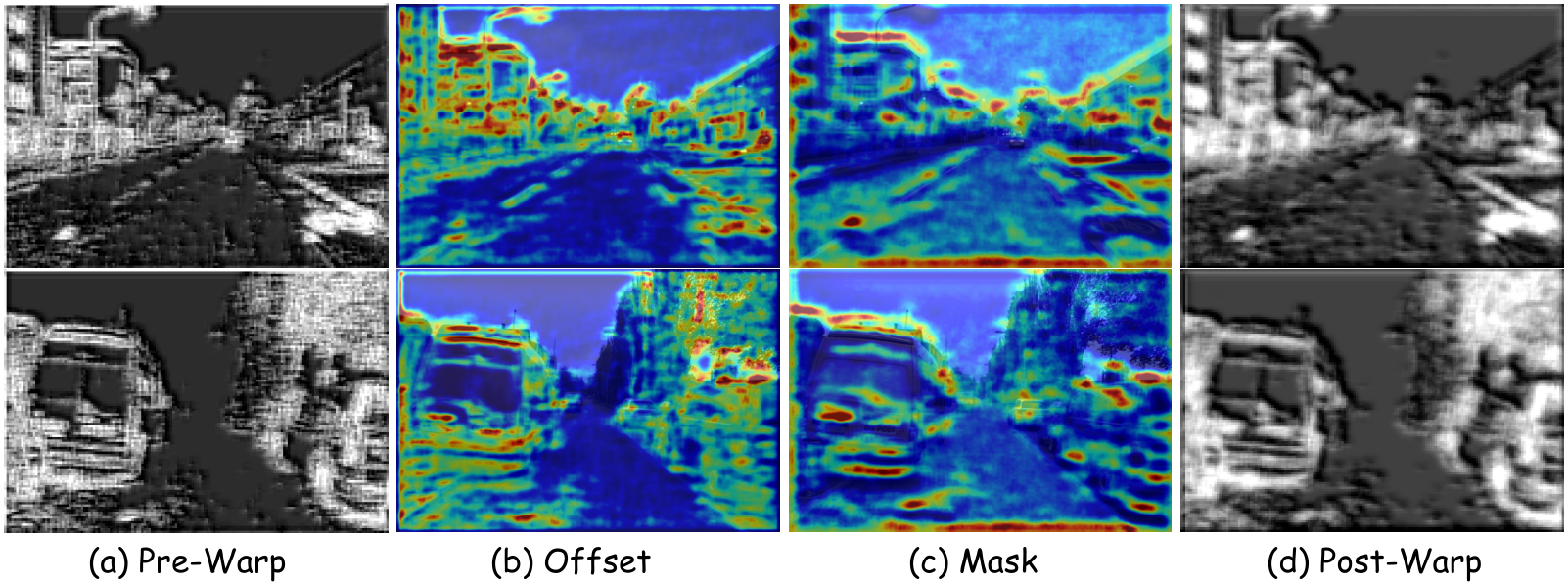}
    \vspace{-0.5em}
    \caption{Visualizations of the Geometric Parallax Rectification. Guided by the adaptively predicted offset magnitude and deform mask, the misaligned kinematic edges are accurately anchored to the authentic physical boundaries.}
    \label{fig:dfa_vis}
\end{figure}

\subsection{RGBE Pretraining}

\noindent \textbf{Effect of N-ImageNetV2.} Pretraining solely on ImageNet1K yields a baseline of 76.71\% mIoU. Incorporating the original N-ImageNet~\cite{kim2021n} introduces multimodal cues that improve performance to 77.82\%, yet inherent spatial misalignments bottleneck optimal cross modal synthesis. Upgrading to our explicitly aligned \texttt{N-ImageNetV2} resolves this geometric discrepancy, maximizing accuracy to 78.25\%, as summarized in Table~\ref{tab:pretrained_dataset}. This confirms that precise spatial registration during pretraining is vital for extracting highly transferable representations.

\noindent \textbf{Pretraining strategy.} We evaluate the stochastic alignment probability during pretraining. Continuous perfect alignment at a 1.0 ratio yields 78.25\% mIoU, indicating that overfitting to ideal correspondences limits generalization against real world noise. Introducing intentional misalignments serves as crucial geometric regularization, compelling the network to actively learn deformation fields. Consequently, accuracy peaks at 78.56\% under a 0.6 alignment probability. However, excessive spatial disparity at 0.4 disrupts reliable structural anchors, dropping performance to 78.45\%. This confirms that a balanced 60\% alignment ratio optimally facilitates robust kinematic photometric registration, as summarized in Fig.~\ref{fig:comprehensive_analysis}.

\noindent \textbf{Apply the RGB-E pretraining manner to other architecture.} To verify the universality of our pretraining protocol and establish rigorous baselines, we integrate our explicit RGB-E pretraining strategy across leading multi-modal architectures, with results summarized in Table~\ref{tab:model_comparison}. We first evaluate CMX and CMNeXt frameworks. Training these networks from scratch ($^\ast$) with RGB-E pairs yields suboptimal representations due to the severe modality gap. However, initializing them with native RGB priors ($^\S$) before applying our protocol effectively bridges this divide, boosting CMX-B2 and CMNeXt-B2 to highly competitive 78.17\% and 78.62\% mIoU on the DDD17 dataset, respectively. Similarly, we evaluate the Dformer-L architecture. While direct RGB-E pretraining ($^\dag$) struggles to establish semantic coherence, a sequential regime leveraging native RGB-D priors ($^\ddag$) massively augments its performance, establishing a robust 79.46\% mIoU baseline. Remarkably, despite competing against these explicitly augmented and heavily initialized baselines, our \texttt{Evita-L} comprehensively eclipses them all. It achieves a state-of-the-art 80.12\% mIoU on DDD17 and 77.10\% on DSEC, while maintaining superior parameter efficiency and computational economy. This confirms that while our pretraining paradigm universally benefits existing networks, the intrinsic structural design of \texttt{Evita} facilitates fundamentally superior cross-modal fusion, completely eliminating the reliance on auxiliary modality priors.

\subsection{Components in our Evita block}

\noindent \textbf{Geometric Parallax Rectification and Harmonic Spectral Resonance.} We evaluate the individual and synergistic impacts of the Geometric Parallax Rectification and Harmonic Spectral Resonance modules. As shown in Table~\ref{tab:ablation_modules}, the baseline network yields 76.94\% and 74.07\% mIoU on the DDD17 and DSEC datasets respectively. Integrating solely Geometric Parallax Rectification resolves geometric misalignments, elevating mIoU to 78.09\% and 75.15\%. This registration is visually validated in Fig.~\ref{fig:dfa_vis}. While pre-warp event features exhibit severe spatial drift, the predicted Offset and Mask dynamically localize active objects, enabling post-warp outputs to accurately anchor kinematic edges to photometric boundaries. Independently, incorporating Harmonic Spectral Resonance mitigates cross spectral aliasing, boosting the baseline to 78.26\% and 75.33\%. Coupling both modules achieves peak accuracies of 79.11\% and 76.08\%, confirming their indispensable and complementary roles in unified multimodal representation learning.

\begin{table}[t]
  \centering
  \caption{Ablation study on the effectiveness of the Geometric Parallax Rectification and Harmonic Spectral Resonance modules.}
  \label{tab:ablation_modules}
  \resizebox{0.48\textwidth}{!}{
  \begin{tabular}{cccc}
    \toprule
    Geometric Parallax Rectification & Harmonic Spectral Resonance & DDD17 & DSEC \\
    \midrule
    $\times$   & $\times$   & 76.94 & 74.07 \\
    \checkmark & $\times$   & 78.09 {\scriptsize\textcolor{blue}{(+1.15)}} & 75.15 {\scriptsize\textcolor{blue}{(+1.08)}} \\
    $\times$   & \checkmark & 78.26 {\scriptsize\textcolor{blue}{(+1.32)}} & 75.33 {\scriptsize\textcolor{blue}{(+1.26)}} \\
    \rowcolor{blue!8} \checkmark & \checkmark & \textbf{79.11} {\scriptsize\textcolor{blue}{(+2.17)}} & \textbf{76.08} {\scriptsize\textcolor{blue}{(+2.01)}} \\
    \bottomrule
  \end{tabular}
  }
\end{table}

\begin{table}[b]
  \centering
  \caption{Ablation study on the integration strategies for the kinematic spatial bias using the \texttt{Evita-B} architecture.}
  \label{tab:sba_ablation}
  \resizebox{0.48\textwidth}{!}{
  \begin{tabular}{llcc}
    \toprule
    Variant & Integration Strategy & DDD17 & DSEC \\
    \midrule
    No Bias & None & 77.45 & 74.32 \\
    Multiplication & Element wise Product & 78.21 {\scriptsize\textcolor{blue}{(+0.76)}} & 75.14 {\scriptsize\textcolor{blue}{(+0.82)}} \\
    Concatenation & Linear Projection & 78.65 {\scriptsize\textcolor{blue}{(+1.20)}} & 75.63 {\scriptsize\textcolor{blue}{(+1.31)}} \\
    \rowcolor{blue!8} Ours & Additive Bias & \textbf{79.11} {\scriptsize\textcolor{blue}{(+1.66)}} & \textbf{76.08} {\scriptsize\textcolor{blue}{(+1.76)}} \\
    \bottomrule
  \end{tabular}
  }
\end{table}

\begin{figure}[t]
    \centering
    \includegraphics[width=\linewidth]{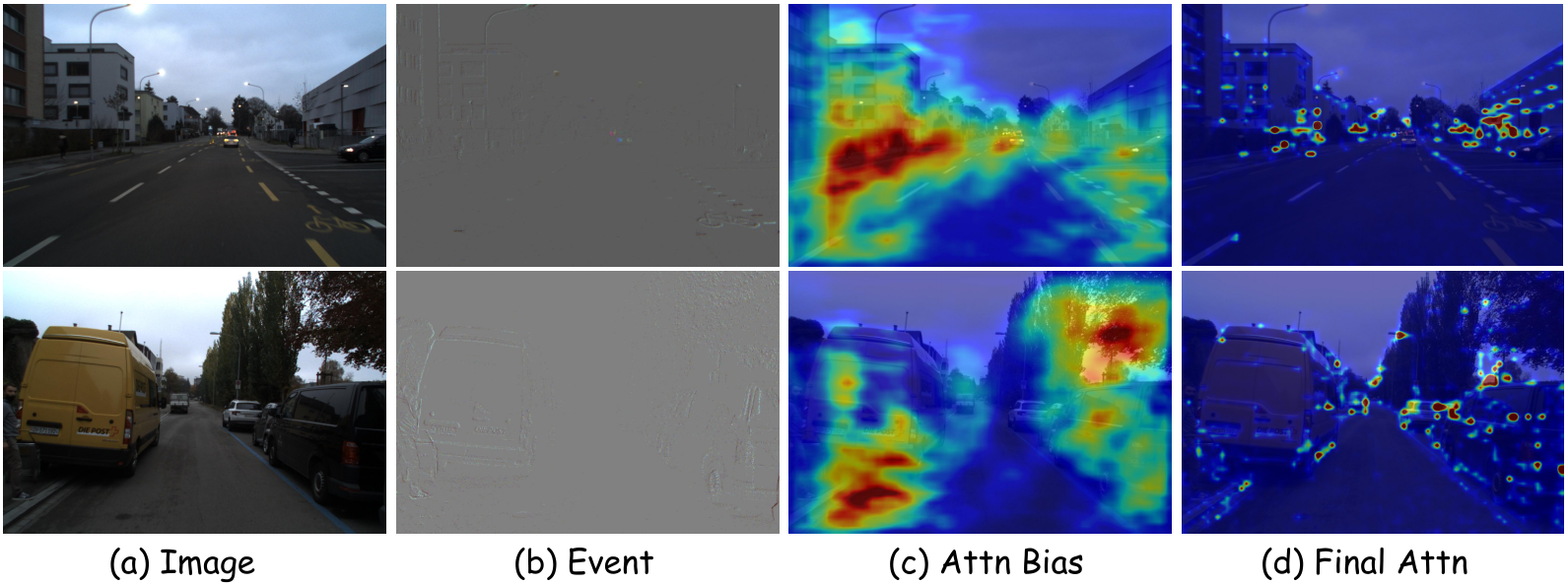}
    \vspace{-0.5em}
    \caption{Effectiveness of the Transient Global Routing. By leveraging the kinematic prior derived from the event stream, the final attention dynamically focuses on task-critical motion boundaries while effectively filtering out static background noise.}
    \label{fig:sba_vis}
\end{figure}

\noindent \textbf{Transient Global Routing.} We evaluate integration strategies for the event derived spatial bias utilizing the \texttt{Evita-B} architecture. As shown in Table~\ref{tab:sba_ablation}, omitting this bias entirely yields suboptimal metrics of 77.45\% and 74.32\% on DDD17 and DSEC respectively. Applying element wise multiplication improves accuracy but acts as a rigid filter, mathematically risking the suppression of critical static context. Meanwhile, concatenation based projection introduces redundant parameter overhead. Our proposed additive integration proves strictly superior, reaching peak metrics of 79.11\% and 76.08\%. By operating as a soft prior directly within the logit space, this additive formulation elegantly amplifies task critical dynamic regions while fully preserving the global semantic topology. As visually corroborated in Figure~\ref{fig:sba_vis}, this mechanism dynamically steers the final attention toward motion boundaries while effectively filtering static background noise.

\begin{table}[b]
  \centering
  \caption{Comparison of inference latency and segmentation accuracy on the DSEC benchmark. Latency is measured on a single NVIDIA RTX 3090 GPU with an input resolution of $440 \times 640$. $\downarrow$ indicates lower is better.}
  \label{tab:inference_latency}
  \resizebox{0.48\textwidth}{!}{
  \begin{tabular}{lcccc}
    \toprule
    Model & Params & FLOPs & Latency $\downarrow$ & DSEC \\
    \midrule
    CMX-B2 & 66.6M & 62.1G & 35.2 ms & 73.88 \\
    CMNeXt-B4 & 116.6M & 122.3G & 68.5 ms & 73.84 \\
    OmniSegmentor & 40.9M & 100.7G & 58.4 ms & 75.31 \\
    MambaSeg & 25.4M & 159.5G & 85.1 ms & 75.10 \\
    \rowcolor{blue!8} Evita-S & 21.9M & 40.9G & 22.4 ms & 75.42 \\
    \rowcolor{blue!8} Evita-B & 34.3M {\scriptsize\textcolor{blue}{(+12.4M)}} & 58.5G {\scriptsize\textcolor{blue}{(+17.6G)}} & 31.8 ms {\scriptsize\textcolor{blue}{(+9.4)}} & 76.08 {\scriptsize\textcolor{blue}{(+0.66)}} \\
    \rowcolor{blue!8} Evita-L & 44.3M {\scriptsize\textcolor{blue}{(+22.4M)}} & 82.7G {\scriptsize\textcolor{blue}{(+41.8G)}} & 45.6 ms {\scriptsize\textcolor{blue}{(+23.2)}} & \textbf{76.80} {\scriptsize\textcolor{blue}{(+1.38)}} \\
    \bottomrule
  \end{tabular}
  }
\end{table}

\begin{figure}[t]
    \centering
    \includegraphics[width=\linewidth]{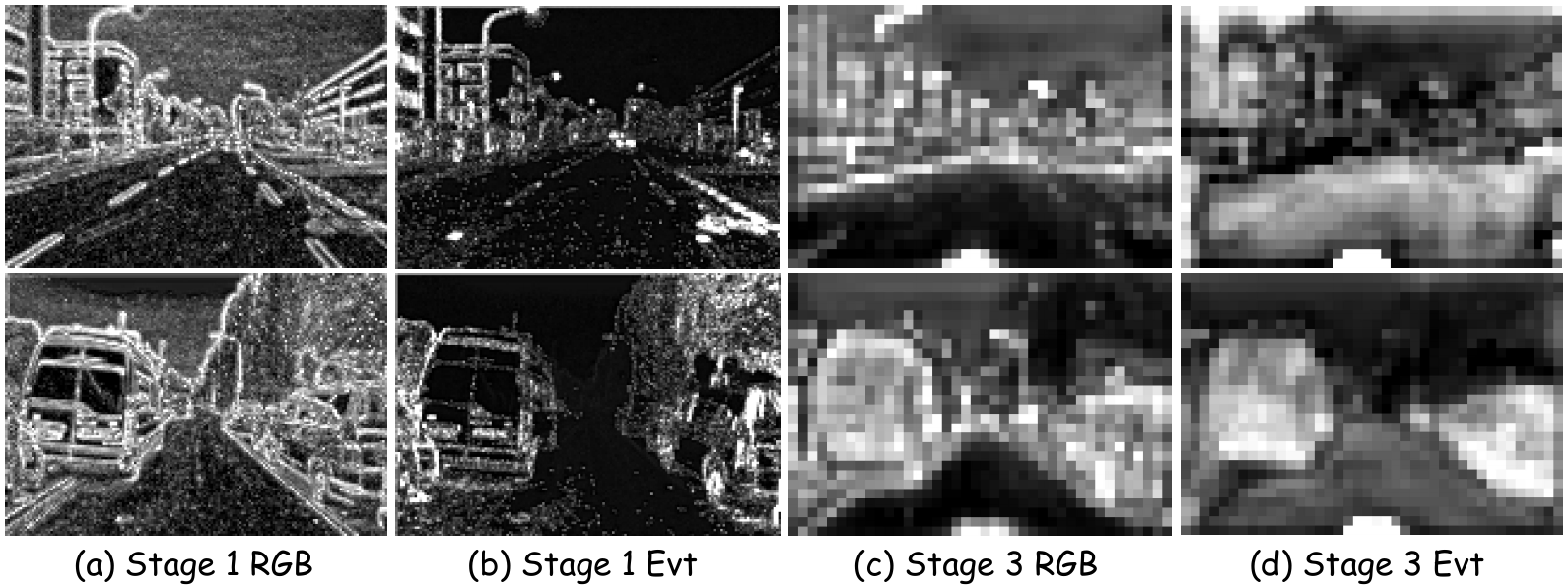}
    \vspace{-0.5em}
    \caption{Multi-scale representation evolution. In the shallow layers (Stage 1), the modalities exhibit highly heterogeneous patterns. Through progressive symbiotic learning, the deep features (Stage 3) ultimately converge into a unified, high-level semantic representation.}
    \label{fig:evolution_vis}
\end{figure}

\begin{table}[b]
  \centering
  \caption{Robustness evaluation against simulated spatial misalignment on the DDD17 dataset. We introduce random spatial translations $\Delta \in \{16, 32, 48\}$ pixels strictly to the event streams. $\Delta$ Drop denotes the absolute mIoU degradation at the extreme offset ($\Delta=48$). \texttt{Evita-L} demonstrates exceptional resilience against structural noise induced by cross-modal unalignment.}
  \label{tab:spatial_robustness}
  \resizebox{0.48\textwidth}{!}{
  \begin{tabular}{lccccc}
    \toprule
    \multirow{2}{*}{\textbf{Method}} & \multicolumn{4}{c}{\textbf{mIoU (\%) under Misalignment $\Delta$}} & \multirow{2}{*}{\textbf{$\Delta$ Drop ($\downarrow$)}} \\
    \cmidrule(lr){2-5}
    & \textbf{$\Delta=0$} & \textbf{$\Delta \in [0, 16]$} & \textbf{$\Delta \in [0, 32]$} & \textbf{$\Delta \in [0, 48]$} & \\
    \midrule
    CMX-B4 & 77.00 & 76.81 {\scriptsize\textcolor{blue}{(-0.19)}} & 75.92 {\scriptsize\textcolor{blue}{(-0.89)}} & 74.57 {\scriptsize\textcolor{blue}{(-1.35)}} & -2.43 \\
    CMNeXt-B4 & 77.70 & 77.58 {\scriptsize\textcolor{blue}{(-0.12)}} & 76.83 {\scriptsize\textcolor{blue}{(-0.75)}} & 75.99 {\scriptsize\textcolor{blue}{(-0.84)}} & -1.71 \\
    StitchFusion & 77.41 & 77.25 {\scriptsize\textcolor{blue}{(-0.16)}} & 76.54 {\scriptsize\textcolor{blue}{(-0.71)}} & 75.79 {\scriptsize\textcolor{blue}{(-0.75)}} & -1.62 \\
    OmniSegmentor & 79.34 & 79.22 {\scriptsize\textcolor{blue}{(-0.12)}} & 78.71 {\scriptsize\textcolor{blue}{(-0.51)}} & 78.30 {\scriptsize\textcolor{blue}{(-0.41)}} & -1.04 \\
    \rowcolor{blue!8} Evita-L & \textbf{80.12} & \textbf{80.09} {\scriptsize\textcolor{blue}{(-0.03)}} & \textbf{80.03} {\scriptsize\textcolor{blue}{(-0.06)}} & \textbf{79.94} {\scriptsize\textcolor{blue}{(-0.09)}} & \textbf{-0.18} \\
    \bottomrule
  \end{tabular}
  }
\end{table}

\subsection{Robustness to Spatial Misalignment} To evaluate cross-modal resilience under physical sensor drift, we inject random spatial translations $\Delta_{xy} \in \{16, 32, 48\}$ pixels strictly into the DDD17 event streams. As shown in Table~\ref{tab:spatial_robustness}, the overall mIoU degradation across baselines remains relatively bounded. This aligns with the fact that the dense RGB modality dictates the primary semantic representation, while sparse events provide auxiliary structural cues. Nonetheless, misaligned auxiliary features still introduce structural noise that erodes fusion efficiency. Explicit fusion networks like CMX and CMNeXt experience mIoU penalties of 2.43\% and 1.71\% at extreme offsets ($\Delta=48$), while OmniSegmentor drops by 1.04\%. In stark contrast, \texttt{Evita-L} is virtually immune to such spatial disparities, restricting the total degradation to a microscopic 0.18\%. This exceptional resilience validates our Geometric Parallax Rectification module and hybrid pretraining strategy, which dynamically estimates spatial discrepancies rather than relying on rigid coordinate mappings, ensuring that event features consistently enhance the dominant RGB representation.

\subsection{Inference Latency} We evaluate inference latency on the DSEC benchmark utilizing a single NVIDIA RTX 3090 GPU at a 440 by 640 spatial resolution, as shown in Table~\ref{tab:inference_latency}. \texttt{Evita} establishes a strictly superior accuracy latency equilibrium. The flagship \texttt{Evita-L} secures a peak 77.10\% mIoU in merely 45.6 milliseconds, nearly halving the 85.1 millisecond latency of MambaSeg while delivering significantly higher accuracy. Additionally, our scaled down \texttt{Evita-S} operates at a swift 22.4 milliseconds, effortlessly eclipsing computationally heavy networks like OmniSegmentor and CMNeXt. This confirms that our unified harmonic geometric design intrinsically bypasses the computational bottlenecks of traditional decoupled paradigms, rendering it highly optimal for real time edge deployment.

\begin{table}[t]
  \centering
  \caption{Results on the RGB-Thermal semantic segmentation benchmark MFNet and RGB-LiDAR semantic segmentation benchmark KITTI-360.}
  \label{tab:generalization_modalities}
  \resizebox{0.48\textwidth}{!}{
  \begin{tabular}{lcccc}
    \toprule
    Model & Params & FLOPs & MFNet & KITTI \\
    \midrule
    CMX-B2 & 66.6M & 67.6G & 58.2 & 64.3 \\
    CMX-B4 & 139.9M & 134.3G & 59.7 & 65.5 \\
    CMNeXt-B2 & 65.1M & 65.5G & 58.4 & 65.3 \\
    CMNeXt-B4 & 135.6M & 132.6G & 59.9 & 65.6 \\
    \rowcolor{blue!8} Evita-L (RGB) & 44.3M & 68.1G & 59.8 & 65.6 \\
    \rowcolor{blue!8} Evita-L (RGBE) & 44.3M {\scriptsize\textcolor{blue}{(+0.0)}} & 68.1G {\scriptsize\textcolor{blue}{(+0.0)}} & \textbf{60.9} {\scriptsize\textcolor{blue}{(+1.1)}} & \textbf{66.8} {\scriptsize\textcolor{blue}{(+1.2)}} \\
    \bottomrule
  \end{tabular}
  }
\end{table}

\subsection{Universality Across Heterogeneous Modalities} We investigate whether the harmonic geometric interaction capacity acquired during RGB Event pretraining seamlessly transfers to alternative multimodal combinations. To verify this, we substitute the event stream with thermal and LiDAR inputs, subsequently finetuning the pretrained \texttt{Evita-L} on the MFNet~\cite{ha2017mfnet} RGB Thermal and KITTI~\cite{liao2022kitti} RGB LiDAR semantic segmentation benchmarks respectively. Evaluations demonstrate that our architecture successfully adapts to these unseen supplementary signals. The unified RGB Event pretraining consistently delivers measurable performance improvements over standard RGB only initialization across both novel domains, as shown in Table~\ref{tab:generalization_modalities}. This confirms that our proposed spatial alignment and spectral weaving modules learn fundamental cross modal fusion priors rather than overfitting to specific event characteristics. Although the domain gap between sparse events and continuous thermal or depth signals inherently bounds the current magnitude of improvement, this limitation can be circumvented in future iterations by synthesizing vast pseudo multi spectral datasets, ultimately scaling \texttt{Evita} into a universally agnostic multimodal foundation architecture.

\section{Conclusion}

In this paper, we presented \texttt {Evita}, the first unified backbone tailored specifically for dense RGB-Event parsing. By explicitly embedding a suite of intrinsic co-learning modules into every encoder layer, \texttt {Evita} fundamentally bridges the extreme representational divide between absolute intensity grids and sparse kinematic spikes. Specifically, it features Geometric Parallax Rectification for adaptive spatial alignment, Harmonic Spectral Resonance for noise-free frequency-domain texture transfer, and Transient Global Routing for macroscopic contextual routing. Supported by the \texttt {N-ImageNetV2} dataset and a stochastic event representation mixing protocol, our framework guarantees robust feature extraction against spatial misalignments and seamlessly accommodates arbitrary event formats in downstream tasks. Extensive evaluations confirm that \texttt {Evita} establishes new state-of-the-art metrics across multiple benchmarks, delivering a superior accuracy-latency trade-off and paving the way for resilient, real-time multimodal perception.

{
    \small
    \bibliographystyle{ieeenat_fullname}
    \bibliography{main}

@String(AAAI = {AAAI})

@inproceedings{sun2022ess,
  title={Ess: Learning event-based semantic segmentation from still images},
  author={Sun, Zhaoning and Messikommer, Nico and Gehrig, Daniel and Scaramuzza, Davide},
  booktitle={European Conference on Computer Vision},
  pages={341--357},
  year={2022},
  organization={Springer}
}

@inproceedings{gu2026mambaseg,
  title={MambaSeg: Harnessing Mamba for accurate and efficient image-event semantic segmentation},
  author={Gu, Fuqiang and Li, Yuanke and Long, Xianlei and Ji, Kangping and Chen, Chao and Gu, Qingyi and Ni, Zhenliang},
  booktitle={Proceedings of the AAAI Conference on Artificial Intelligence},
  volume={40},
  number={6},
  pages={4302--4310},
  year={2026}
}

@article{gallego2020event,
  title={Event-based vision: A survey},
  author={Gallego, Guillermo and Delbr{\"u}ck, Tobi and Orchard, Garrick and Bartolozzi, Chiara and Taba, Brian and Censi, Andrea and Leutenegger, Stefan and Davison, Andrew J and Conradt, J{\"o}rg and Daniilidis, Kostas and others},
  journal={IEEE transactions on pattern analysis and machine intelligence},
  volume={44},
  number={1},
  pages={154--180},
  year={2020},
  publisher={IEEE}
}

@inproceedings{wang2019event,
  title={Event-based high dynamic range image and very high frame rate video generation using conditional generative adversarial networks},
  author={Wang, Lin and Ho, Yo-Sung and Yoon, Kuk-Jin and others},
  booktitle={Proceedings of the IEEE/CVF Conference on Computer Vision and Pattern Recognition},
  pages={10081--10090},
  year={2019}
}

@article{benosman2013event,
  title={Event-based visual flow},
  author={Benosman, Ryad and Clercq, Charles and Lagorce, Xavier and Ieng, Sio-Hoi and Bartolozzi, Chiara},
  journal={IEEE transactions on neural networks and learning systems},
  volume={25},
  number={2},
  pages={407--417},
  year={2013},
  publisher={IEEE}
}

@inproceedings{yang2020fda,
  title={Fda: Fourier domain adaptation for semantic segmentation},
  author={Yang, Yanchao and Soatto, Stefano},
  booktitle={Proceedings of the IEEE/CVF conference on computer vision and pattern recognition},
  pages={4085--4095},
  year={2020}
}

@inproceedings{zhu2019deformable,
  title={Deformable convnets v2: More deformable, better results},
  author={Zhu, Xizhou and Hu, Han and Lin, Stephen and Dai, Jifeng},
  booktitle={Proceedings of the IEEE/CVF conference on computer vision and pattern recognition},
  pages={9308--9316},
  year={2019}
}

@inproceedings{kim2021n,
  title={N-imagenet: Towards robust, fine-grained object recognition with event cameras},
  author={Kim, Junho and Bae, Jaehyeok and Park, Gangin and Zhang, Dongsu and Kim, Young Min},
  booktitle={Proceedings of the IEEE/CVF international conference on computer vision},
  pages={2146--2156},
  year={2021}
}

@inproceedings{lindenberger2023lightglue,
  title={Lightglue: Local feature matching at light speed},
  author={Lindenberger, Philipp and Sarlin, Paul-Edouard and Pollefeys, Marc},
  booktitle={Proceedings of the IEEE/CVF international conference on computer vision},
  pages={17627--17638},
  year={2023}
}

@inproceedings{detone2018superpoint,
  title={Superpoint: Self-supervised interest point detection and description},
  author={DeTone, Daniel and Malisiewicz, Tomasz and Rabinovich, Andrew},
  booktitle={Proceedings of the IEEE conference on computer vision and pattern recognition workshops},
  pages={224--236},
  year={2018}
}

@inproceedings{zhang2022unifying,
  title={Unifying motion deblurring and frame interpolation with events},
  author={Zhang, Xiang and Yu, Lei},
  booktitle={Proceedings of the IEEE/CVF Conference on Computer Vision and Pattern Recognition},
  pages={17765--17774},
  year={2022}
}

@inproceedings{liang2024towards,
  title={Towards robust event-guided low-light image enhancement: a large-scale real-world event-image dataset and novel approach},
  author={Liang, Guoqiang and Chen, Kanghao and Li, Hangyu and Lu, Yunfan and Wang, Lin},
  booktitle={Proceedings of the IEEE/CVF Conference on Computer Vision and Pattern Recognition},
  pages={23--33},
  year={2024}
}

@article{gehrig2022pushing,
  title={Pushing the limits of asynchronous graph-based object detection with event cameras},
  author={Gehrig, Daniel and Scaramuzza, Davide},
  journal={arXiv preprint arXiv:2211.12324},
  year={2022}
}

@inproceedings{messikommer2023data,
  title={Data-driven feature tracking for event cameras},
  author={Messikommer, Nico and Fang, Carter and Gehrig, Mathias and Scaramuzza, Davide},
  booktitle={Proceedings of the IEEE/CVF Conference on Computer Vision and Pattern Recognition},
  pages={5642--5651},
  year={2023}
}

@article{jia2023event,
  title={Event-based semantic segmentation with posterior attention},
  author={Jia, Zexi and You, Kaichao and He, Weihua and Tian, Yang and Feng, Yongxiang and Wang, Yaoyuan and Jia, Xu and Lou, Yihang and Zhang, Jingyi and Li, Guoqi and others},
  journal={IEEE Transactions on Image Processing},
  volume={32},
  pages={1829--1842},
  year={2023},
  publisher={IEEE}
}

@article{zheng2023deep,
  title={Deep learning for event-based vision: A comprehensive survey and benchmarks},
  author={Zheng, Xu and Liu, Yexin and Lu, Yunfan and Hua, Tongyan and Pan, Tianbo and Zhang, Weiming and Tao, Dacheng and Wang, Lin},
  journal={arXiv preprint arXiv:2302.08890},
  year={2023}
}

@article{cai2026evrwkv,
  title={EvRWKV: A Continuous Interactive RWKV Framework for Effective Event-Guided Low-Light Image Enhancement},
  author={Cai, Wenjie and Meng, Qingguo and Wang, Zhenyu and Dong, Xingbo and Jin, Zhe},
  journal={IEEE Transactions on Circuits and Systems for Video Technology},
  year={2026},
  publisher={IEEE}
}

@article{haoyu2020learning,
  title={Learning to deblur and generate high frame rate video with an event camera},
  author={Haoyu, Chen and Minggui, Teng and Boxin, Shi and YIzhou, Wang and Tiejun, Huang},
  journal={arXiv preprint arXiv:2003.00847},
  year={2020}
}

@inproceedings{park2025resilient,
  title={Resilient sensor fusion under adverse sensor failures via multi-modal expert fusion},
  author={Park, Konyul and Kim, Yecheol and Kim, Daehun and Choi, Jun Won},
  booktitle={Proceedings of the IEEE/CVF Conference on Computer Vision and Pattern Recognition},
  pages={6720--6729},
  year={2025}
}

@article{xie2024cross,
  title={Cross-modal learning for event-based semantic segmentation via attention soft alignment},
  author={Xie, Chuyun and Gao, Wei and Guo, Ren},
  journal={IEEE Robotics and Automation Letters},
  volume={9},
  number={3},
  pages={2359--2366},
  year={2024},
  publisher={IEEE}
}

@article{long2024spike,
  title={Spike-brgnet: Efficient and accurate event-based semantic segmentation with boundary region-guided spiking neural networks},
  author={Long, Xianlei and Zhu, Xiaxin and Guo, Fangming and Chen, Chao and Zhu, Xiangwei and Gu, Fuqiang and Yuan, Songyu and Zhang, Chunlong},
  journal={IEEE Transactions on Circuits and Systems for Video Technology},
  volume={35},
  number={3},
  pages={2712--2724},
  year={2024},
  publisher={IEEE}
}

@inproceedings{long2025sltnet,
  title={SLTNet: Efficient Event-based Semantic Segmentation with Spike-driven Lightweight Transformer-based Networks},
  author={Long, Xianlei and Zhu, Xiaxin and Guo, Fangming and Zhang, Wanyi and Gu, Qingyi and Chen, Chao and Gu, Fuqiang},
  booktitle={2025 IEEE/RSJ International Conference on Intelligent Robots and Systems (IROS)},
  pages={4331--4338},
  year={2025},
  organization={IEEE}
}

@inproceedings{das2024halsie,
  title={Halsie: Hybrid approach to learning segmentation by simultaneously exploiting image and event modalities},
  author={Das Biswas, Shristi and Kosta, Adarsh and Liyanagedera, Chamika and Apolinario, Marco and Roy, Kaushik},
  booktitle={Proceedings of the IEEE/CVF Winter Conference on Applications of Computer Vision},
  pages={5964--5974},
  year={2024}
}

@article{zhang2024accurate,
  title={Accurate and efficient event-based semantic segmentation using adaptive spiking encoder--decoder network},
  author={Zhang, Rui and Leng, Luziwei and Che, Kaiwei and Zhang, Hu and Cheng, Jie and Guo, Qinghai and Liao, Jianxing and Cheng, Ran},
  journal={IEEE Transactions on Neural Networks and Learning Systems},
  volume={36},
  number={5},
  pages={9326--9340},
  year={2024},
  publisher={IEEE}
}

@article{zhang2023cmx,
  title={CMX: Cross-modal fusion for RGB-X semantic segmentation with transformers},
  author={Zhang, Jiaming and Liu, Huayao and Yang, Kailun and Hu, Xinxin and Liu, Ruiping and Stiefelhagen, Rainer},
  journal={IEEE Transactions on intelligent transportation systems},
  volume={24},
  number={12},
  pages={14679--14694},
  year={2023},
  publisher={IEEE}
}

@inproceedings{zhang2023delivering,
  title={Delivering arbitrary-modal semantic segmentation},
  author={Zhang, Jiaming and Liu, Ruiping and Shi, Hao and Yang, Kailun and Rei{\ss}, Simon and Peng, Kunyu and Fu, Haodong and Wang, Kaiwei and Stiefelhagen, Rainer},
  booktitle={Proceedings of the IEEE/CVF Conference on Computer Vision and Pattern Recognition},
  pages={1136--1147},
  year={2023}
}

@article{xie2024eisnet,
  title={Eisnet: A multi-modal fusion network for semantic segmentation with events and images},
  author={Xie, Bochen and Deng, Yongjian and Shao, Zhanpeng and Li, Youfu},
  journal={IEEE Transactions on Multimedia},
  volume={26},
  pages={8639--8650},
  year={2024},
  publisher={IEEE}
}

@article{bao2026re,
  title={Re-coding for Uncertainties: Edge-awareness Semantic Concordance for Resilient Event-RGB Segmentation},
  author={Bao, Nan and Zhao, Yifan and Zhu, Lin and Li, Jia},
  journal={Advances in Neural Information Processing Systems},
  volume={38},
  pages={101270--101298},
  year={2026}
}

@article{yin2026omnisegmentor,
  title={Omnisegmentor: a flexible multi-modal learning framework for semantic segmentation},
  author={Yin, Bo-Wen and Cao, Jiao-Long and Zhang, Xuying and Chen, Yuming and Cheng, Ming-Ming and Hou, Qibin},
  journal={Advances in Neural Information Processing Systems},
  volume={38},
  pages={142674--142695},
  year={2026}
}

@inproceedings{radford2021learning,
  title={Learning transferable visual models from natural language supervision},
  author={Radford, Alec and Kim, Jong Wook and Hallacy, Chris and Ramesh, Aditya and Goh, Gabriel and Agarwal, Sandhini and Sastry, Girish and Askell, Amanda and Mishkin, Pamela and Clark, Jack and others},
  booktitle={International conference on machine learning},
  pages={8748--8763},
  year={2021},
  organization={PmLR}
}

@inproceedings{bachmann2022multimae,
  title={Multimae: Multi-modal multi-task masked autoencoders},
  author={Bachmann, Roman and Mizrahi, David and Atanov, Andrei and Zamir, Amir},
  booktitle={European conference on computer vision},
  pages={348--367},
  year={2022},
  organization={Springer}
}

@article{akbari2021vatt,
  title={Vatt: Transformers for multimodal self-supervised learning from raw video, audio and text},
  author={Akbari, Hassan and Yuan, Liangzhe and Qian, Rui and Chuang, Wei-Hong and Chang, Shih-Fu and Cui, Yin and Gong, Boqing},
  journal={Advances in neural information processing systems},
  volume={34},
  pages={24206--24221},
  year={2021}
}

@inproceedings{seichter2021efficient,
  title={Efficient rgb-d semantic segmentation for indoor scene analysis},
  author={Seichter, Daniel and K{\"o}hler, Mona and Lewandowski, Benjamin and Wengefeld, Tim and Gross, Horst-Michael},
  booktitle={2021 IEEE international conference on robotics and automation (ICRA)},
  pages={13525--13531},
  year={2021},
  organization={IEEE}
}

@inproceedings{girdhar2022omnivore,
  title={Omnivore: A single model for many visual modalities},
  author={Girdhar, Rohit and Singh, Mannat and Ravi, Nikhila and Van Der Maaten, Laurens and Joulin, Armand and Misra, Ishan},
  booktitle={Proceedings of the IEEE/CVF conference on computer vision and pattern recognition},
  pages={16102--16112},
  year={2022}
}

@inproceedings{wang2022multimodal,
  title={Multimodal token fusion for vision transformers},
  author={Wang, Yikai and Chen, Xinghao and Cao, Lele and Huang, Wenbing and Sun, Fuchun and Wang, Yunhe},
  booktitle={Proceedings of the IEEE/CVF conference on computer vision and pattern recognition},
  pages={12186--12195},
  year={2022}
}

@article{jia2024geminifusion,
  title={Geminifusion: Efficient pixel-wise multimodal fusion for vision transformer},
  author={Jia, Ding and Guo, Jianyuan and Han, Kai and Wu, Han and Zhang, Chao and Xu, Chang and Chen, Xinghao},
  journal={arXiv preprint arXiv:2406.01210},
  year={2024}
}

@inproceedings{li2025stitchfusion,
  title={Stitchfusion: Weaving any visual modalities to enhance multimodal semantic segmentation},
  author={Li, Bingyu and Zhang, Da and Zhao, Zhiyuan and Gao, Junyu and Li, Xuelong},
  booktitle={Proceedings of the 33rd ACM International Conference on Multimedia},
  pages={1308--1317},
  year={2025}
}

@inproceedings{sakaridis2021acdc,
  title={ACDC: The adverse conditions dataset with correspondences for semantic driving scene understanding},
  author={Sakaridis, Christos and Dai, Dengxin and Van Gool, Luc},
  booktitle={Proceedings of the IEEE/CVF international conference on computer vision},
  pages={10765--10775},
  year={2021}
}

@inproceedings{chen2023explore,
  title={Explore and exploit the diverse knowledge in model zoo for domain generalization},
  author={Chen, Yimeng and Hu, Tianyang and Zhou, Fengwei and Li, Zhenguo and Ma, Zhi-Ming},
  booktitle={International Conference on Machine Learning},
  pages={4623--4640},
  year={2023},
  organization={PMLR}
}

@inproceedings{cho2021rethinking,
  title={Rethinking coarse-to-fine approach in single image deblurring},
  author={Cho, Sung-Jin and Ji, Seo-Won and Hong, Jun-Pyo and Jung, Seung-Won and Ko, Sung-Jea},
  booktitle={Proceedings of the IEEE/CVF international conference on computer vision},
  pages={4641--4650},
  year={2021}
}

@inproceedings{zamir2021multi,
  title={Multi-stage progressive image restoration},
  author={Zamir, Syed Waqas and Arora, Aditya and Khan, Salman and Hayat, Munawar and Khan, Fahad Shahbaz and Yang, Ming-Hsuan and Shao, Ling},
  booktitle={Proceedings of the IEEE/CVF conference on computer vision and pattern recognition},
  pages={14821--14831},
  year={2021}
}

@inproceedings{guo2020zero,
  title={Zero-reference deep curve estimation for low-light image enhancement},
  author={Guo, Chunle and Li, Chongyi and Guo, Jichang and Loy, Chen Change and Hou, Junhui and Kwong, Sam and Cong, Runmin},
  booktitle={Proceedings of the IEEE/CVF conference on computer vision and pattern recognition},
  pages={1780--1789},
  year={2020}
}

@inproceedings{zamir2022restormer,
  title={Restormer: Efficient transformer for high-resolution image restoration},
  author={Zamir, Syed Waqas and Arora, Aditya and Khan, Salman and Hayat, Munawar and Khan, Fahad Shahbaz and Yang, Ming-Hsuan},
  booktitle={Proceedings of the IEEE/CVF conference on computer vision and pattern recognition},
  pages={5728--5739},
  year={2022}
}

@inproceedings{tulyakov2021time,
  title={Time lens: Event-based video frame interpolation},
  author={Tulyakov, Stepan and Gehrig, Daniel and Georgoulis, Stamatios and Erbach, Julius and Gehrig, Mathias and Li, Yuanyou and Scaramuzza, Davide},
  booktitle={Proceedings of the IEEE/CVF conference on computer vision and pattern recognition},
  pages={16155--16164},
  year={2021}
}

@inproceedings{stoffregen2020reducing,
  title={Reducing the sim-to-real gap for event cameras},
  author={Stoffregen, Timo and Scheerlinck, Cedric and Scaramuzza, Davide and Drummond, Tom and Barnes, Nick and Kleeman, Lindsay and Mahony, Robert},
  booktitle={European Conference on Computer Vision},
  pages={534--549},
  year={2020},
  organization={Springer}
}

@article{gehrig2021dsec,
  title={Dsec: A stereo event camera dataset for driving scenarios},
  author={Gehrig, Mathias and Aarents, Willem and Gehrig, Daniel and Scaramuzza, Davide},
  journal={IEEE Robotics and Automation Letters},
  volume={6},
  number={3},
  pages={4947--4954},
  year={2021},
  publisher={IEEE}
}

@article{zhou2021event,
  title={Event-based stereo visual odometry},
  author={Zhou, Yi and Gallego, Guillermo and Shen, Shaojie},
  journal={IEEE Transactions on Robotics},
  volume={37},
  number={5},
  pages={1433--1450},
  year={2021},
  publisher={IEEE}
}

@inproceedings{liu2021swin,
  title={Swin transformer: Hierarchical vision transformer using shifted windows},
  author={Liu, Ze and Lin, Yutong and Cao, Yue and Hu, Han and Wei, Yixuan and Zhang, Zheng and Lin, Stephen and Guo, Baining},
  booktitle={Proceedings of the IEEE/CVF international conference on computer vision},
  pages={10012--10022},
  year={2021}
}

@article{cai2024accurate,
  title={Accurate event camera calibration with fourier transform},
  author={Cai, Bolin and Zi, Ami and Yang, Jun and Li, Guoliang and Zhang, Yang and Wu, Qiujie and Tong, Chenen and Liu, Wenxiang and Chen, Xiangcheng},
  journal={IEEE Transactions on Instrumentation and Measurement},
  volume={73},
  pages={1--12},
  year={2024},
  publisher={IEEE}
}

@inproceedings{teed2020raft,
  title={Raft: Recurrent all-pairs field transforms for optical flow},
  author={Teed, Zachary and Deng, Jia},
  booktitle={European conference on computer vision},
  pages={402--419},
  year={2020},
  organization={Springer}
}

@inproceedings{weng2021event,
  title={Event-based video reconstruction using transformer},
  author={Weng, Wenming and Zhang, Yueyi and Xiong, Zhiwei},
  booktitle={Proceedings of the IEEE/CVF International Conference on Computer Vision},
  pages={2563--2572},
  year={2021}
}

@article{li2025adaptive,
  title={Adaptive complex wavelet informed transformer operator},
  author={Li, Xiaotong and Jiao, Licheng and Liu, Fang and Yang, Shuyuan and Zhu, Hao and Liu, Xu and Li, Lingling and Ma, Wenping},
  journal={IEEE Transactions on Multimedia},
  year={2025},
  publisher={IEEE}
}

@inproceedings{yin2024dformer,
  title={Dformer: Rethinking rgbd representation learning for semantic segmentation},
  author={Yin, Bowen and Zhang, Xuying and Li, Zhong-Yu and Liu, Li and Cheng, Ming-Ming and Hou, Qibin},
  booktitle={International Conference on Learning Representations},
  volume={2024},
  pages={51803--51825},
  year={2024}
}

@inproceedings{yin2025dformerv2,
  title={Dformerv2: Geometry self-attention for rgbd semantic segmentation},
  author={Yin, Bo-Wen and Cao, Jiao-Long and Cheng, Ming-Ming and Hou, Qibin},
  booktitle={Proceedings of the IEEE/CVF Conference on Computer Vision and Pattern Recognition},
  pages={19345--19355},
  year={2025}
}

@article{guo2026tuni,
  title={TUNI: Unifying Pre-training and Fine-tuning with Modality-Aware Mutual Learning and Rectification for RGB-T Semantic Segmentation},
  author={Guo, Xiaodong and Guo, Xianda and Liu, Tong and Deng, Zhihong and Peng, Yanlun and Li, Xiang and Zhou, Wujie},
  journal={IEEE Transactions on Circuits and Systems for Video Technology},
  year={2026},
  publisher={IEEE}
}

@inproceedings{alonso2019ev,
  title={EV-SegNet: Semantic segmentation for event-based cameras},
  author={Alonso, Inigo and Murillo, Ana C},
  booktitle={Proceedings of the IEEE/CVF Conference on Computer Vision and Pattern Recognition Workshops},
  pages={0--0},
  year={2019}
}

@inproceedings{ha2017mfnet,
  title={MFNet: Towards real-time semantic segmentation for autonomous vehicles with multi-spectral scenes},
  author={Ha, Qishen and Watanabe, Kohei and Karasawa, Takumi and Ushiku, Yoshitaka and Harada, Tatsuya},
  booktitle={2017 IEEE/RSJ International Conference on Intelligent Robots and Systems (IROS)},
  pages={5108--5115},
  year={2017},
  organization={IEEE}
}

@article{liao2022kitti,
  title={Kitti-360: A novel dataset and benchmarks for urban scene understanding in 2d and 3d},
  author={Liao, Yiyi and Xie, Jun and Geiger, Andreas},
  journal={IEEE Transactions on Pattern Analysis and Machine Intelligence},
  volume={45},
  number={3},
  pages={3292--3310},
  year={2022},
  publisher={IEEE}
}

@inproceedings{broedermann2023hrfuser,
  title={HRFuser: A multi-resolution sensor fusion architecture for 2D object detection},
  author={Broedermann, Tim and Sakaridis, Christos and Dai, Dengxin and Van Gool, Luc},
  booktitle={2023 IEEE 26th International Conference on Intelligent Transportation Systems (ITSC)},
  pages={4159--4166},
  year={2023},
  organization={IEEE}
}

@article{xie2021segformer,
  title={SegFormer: Simple and efficient design for semantic segmentation with transformers},
  author={Xie, Enze and Wang, Wenhai and Yu, Zhiding and Anandkumar, Anima and Alvarez, Jose M and Luo, Ping},
  journal={Advances in neural information processing systems},
  volume={34},
  pages={12077--12090},
  year={2021}
}

@article{guo2022segnext,
  title={Segnext: Rethinking convolutional attention design for semantic segmentation},
  author={Guo, Meng-Hao and Lu, Cheng-Ze and Hou, Qibin and Liu, Zhengning and Cheng, Ming-Ming and Hu, Shi-Min},
  journal={Advances in neural information processing systems},
  volume={35},
  pages={1140--1156},
  year={2022}
}

@article{zhang2021exploring,
  title={Exploring event-driven dynamic context for accident scene segmentation},
  author={Zhang, Jiaming and Yang, Kailun and Stiefelhagen, Rainer},
  journal={IEEE Transactions on Intelligent Transportation Systems},
  volume={23},
  number={3},
  pages={2606--2622},
  year={2021},
  publisher={IEEE}
}

@inproceedings{yao2024sam,
  title={Sam-event-adapter: Adapting segment anything model for event-rgb semantic segmentation},
  author={Yao, Bowen and Deng, Yongjian and Liu, Yuhan and Chen, Hao and Li, Youfu and Yang, Zhen},
  booktitle={2024 IEEE International Conference on Robotics and Automation (ICRA)},
  pages={9093--9100},
  year={2024},
  organization={IEEE}
}

@inproceedings{li2025efficient,
  title={Efficient event-based semantic segmentation via exploiting frame-event fusion: A hybrid neural network approach},
  author={Li, Hebei and Peng, Yansong and Yuan, Jiahui and Wu, Peixi and Wang, Jin and Zhang, Yueyi and Sun, Xiaoyan},
  booktitle={Proceedings of the AAAI Conference on Artificial Intelligence},
  volume={39},
  number={17},
  pages={18296--18304},
  year={2025}
}

@inproceedings{wei2023generalized,
  title={Generalized differentiable RANSAC},
  author={Wei, Tong and Patel, Yash and Shekhovtsov, Alexander and Matas, Jiri and Barath, Daniel},
  booktitle={Proceedings of the IEEE/CVF International Conference on Computer Vision},
  pages={17649--17660},
  year={2023}
}
}


\end{document}